\documentclass[twoside,11pt]{article}

\usepackage[abbrvbib, preprint]{jmlr2e}

\usepackage[utf8]{inputenc} 
\usepackage[T1]{fontenc}    
\usepackage{hyperref}       
\usepackage{url}            
\usepackage{booktabs}       
\usepackage{amsfonts}       
\usepackage{nicefrac}       
\usepackage{microtype}      
\usepackage{helvet}  
\usepackage{courier}  
\usepackage{url}  
\usepackage{graphicx}  

\usepackage{algorithmic}
\usepackage{algorithm}
\usepackage{wrapfig}
\usepackage{lipsum}

\usepackage{times}
\usepackage{xspace}

\usepackage{graphicx} 
\usepackage{subfigure} 
\usepackage{caption}
\usepackage{paralist}
\usepackage{float}
\usepackage{tabularx}
\usepackage{multirow}
\usepackage{diagbox}
\usepackage{tikz}
\usepackage{booktabs}
\usepackage{array, makecell} 

\usepackage{natbib}
\usepackage{amsmath,amssymb}
\usepackage[inline]{enumitem}
\usepackage{acronym}

\DeclareMathOperator{\E}{\mathbb{E}}
\DeclareMathOperator*{\argmax}{arg\,max}

\newcounter{todocnt}

\newcounter{latercnt}

\acrodef{RL}{Reinforcement Learning}
\acrodef{DRL}{Deep Reinforcement Learning}
\acrodef{IRL}{Inverse Reinforcement Learning}
\acrodef{SERP}{search engine result page}
\acrodef{IR}{Information Retrieval}
\acrodef{MDP}{Markov Decision Process}
\acrodef{MaxEnt-IRL}{Maximum Entropy Inverse Reinforcement Learning}
\acrodef{DM-IRL}{Distance Minimization Inverse Reinforcement Learning}
\acrodef{ISO}{Interactive System Optimizer}
\acrodef{AIRL}{Adversarial Inverse Reinforcement Learning}
\acrodef{PPO}{Proximal Policy Optimization}

\newcommand{\is}{interactive system\xspace}
\newcommand{\iss}{interactive systems\xspace}
\newcommand{\Iss}{Interactive systems\xspace}

\newcommand{\optimize}{Eq.~\ref{eq:optimal_environment}\xspace}

\newcommand{\negskip}{\vspace*{-.5\baselineskip}}

\begin{document}

\title{Optimizing Interactive Systems via Data-Driven Objectives}

\author{ \name Ziming Li   \email z.li@uva.nl \\
         \addr University of Amsterdam\\
          The Netherlands 
          \AND
                  \name  Julia Kiseleva  \email julia.kiseleva@microsoft.com\\
         \addr Microsoft Research \\
          Redmond, USA
          \AND
                    \name  Alekh Agarwal  \email alekha@microsoft.com\\
         \addr Microsoft Research \\
          Redmond, USA
          \AND
                   \name Maarten de Rijke   \email m.derijke@uva.nl\\
         \addr University of Amsterdam\\
          The Netherlands 
          \AND
                   \name  Ryen W. White   \email  ryenw@microsoft.com\\
         \addr Microsoft Research \\
          Redmond, USA}

\editor{AA and BB}
\maketitle

\begin{abstract} 
Effective optimization is essential for real world \iss to provide a satisfactory user experience in response to changing user behavior.
However, it is often challenging to find an objective to optimize for interactive systems (e.g., policy learning in task-oriented dialog systems). 
Generally, such objectives are manually crafted and rarely capture complex user needs in an accurate manner.
We propose an approach that infers the objective directly from observed user interactions.
These inferences can be made regardless of prior knowledge and across different types of user behavior.
We introduce \acf{ISO}, a novel algorithm that uses these inferred objectives for optimization.
Our main contribution is a new general principled approach to optimizing \iss using data-driven objectives.
We demonstrate the high effectiveness of \ac{ISO} over several simulations.
\end{abstract}


\negskip
\section{Introduction}
\label{sec:intro}

\Iss~\citep{white2016interactions} play an important role in assisting people in a wide range of tasks. For instance, if users are seeking information, \iss can assist them in the form of web search engines~\citep{Williams_www_2016, borisov_www_2016, dehghani2017neural, Williams_sigir_2016}, dialog systems~\citep{Li_emnlp_2016,li2019dialogue,dhingra2016towards,williams2017hybrid, peng2018adversarial}, digital assistants~\citep{kiseleva_chiir_2016,kiseleva_sigir_2016,kiseleva2017evaluating,ter2020conversations}, recommender systems~\citep{schnabel2019shaping,sepliarskaia2018preference}, or virtual reality~\citep{argelaguet2016role}. The described instances of \iss can be considered as examples of machine learning applications where the goal is to assist users in real world day-to-day tasks.  
These systems are characterized by repeated interactions with humans which follow the request-response schema, where the user takes an action, followed by a response from the \is. Such interactions can continue for several iterations until the user decides to stop, e.g., when they are either satisfied or frustrated with their experience. Interaction with the system produces traces/trajectories of user interactions. Importantly, an \is and its users always have a shared goal: for users to have the best experience in the premise of successfully completing the user's task.  

Thus, both a system and its users are expected to behave accordingly, e.g., a searcher issues a query that he expects will lead him to the desired results and the \is provides the search results that are most helpful to him.
However, despite their shared goal, only the user can observe their own experience, leaving \iss unable to directly optimize their own behavior. Given as an example, in most cases, users will not leave explicit rates about their experience after interacting with digital assistants and this brings difficulties in optimizing the system.  

Understanding user objectives and acting accordingly has been shown to be a difficult task, even for humans~\citep{perner1999development}. However, studies in behavioral economics provide supporting evidence that users intend to maximize expected utility or minimize expected cost and effort~\citep{varian1999economics, Lovett_choice_theory, blume2008new}. Following this line of research, in this paper, we assume that the general population of users has a shared goal that is achieved through interactions, so user behaviors are aligned with their preferences but rational noises are allowed. We call such behavior \emph{approximately rational}. 
Obviously, the exact user utility function is inherently complex, but we can approximate it via some meaningful \emph{objective function}, recovered directly from observed traces of user interactions. 
A similar principle has been successfully employed in robotics~\citep{jeon2020reward,reddy2019reward}, and understanding user behavior on web~\citep{azzopardi_sigir_2014, Kosinski_2013, wei_wsdm_2017}.
Hence, knowing the \emph{approximate}  user objective function can help us to improve the flow of \iss.

Currently, optimizing \iss relies on explicit assumptions about users' objectives in terms of their needs and frustrations~\citep{li2017towards}.
Commonly, an objective function is manually designed for a particular task to reflect the quality of an \is, e.g., in terms of user satisfaction~\citep{kelly-methods-2009, kelly2015effort}, user effort~\citep{Yilmaz_cikm_2014} or other domain-specific metrics, such as relevance judgements in information retrieval \citep{Kalervo_2002,sara:rele75,sara:stud88,Drutsa_wsdm_2015,Dupret_wsdm_2013}, user feedbacks (e.g. click, order, skip) in recommender systems \citep{zhao2018recommendations,zheng2018drn,chen2019top}. 

The drawbacks of this approach are that a handcrafted objective function is heavily based on domain knowledge, that it is expensive to maintain, and that it does not generalize over different tasks, e.g., clicks on search results, gestures for mobile digital assistants~\citep{kiseleva_sigir_2016, Williams_www_2016}, the cross-entropy between generated replies and predefined answers~\citep{Li_emnlp_2016,cui2019user}. Consequently, manually crafted objective functions rarely correspond to the actual user experience. Therefore, even an \is that maximizes an objective function is not expected to provide an optimal experience as long as that objective function is hand-crafted.
Moreover, it is impossible to design such functions when there is a lack of domain knowledge. Also, we have witnessed how the badly designed objective functions can lead to wrong results. For example, ~\citet{liu-etal-2016-evaluate} validated that applying evaluation metrics (e.g., BLEU score \citep{papineni2002bleu}) in the machine-translation field to dialog systems is problematic because there is significant diversity in the space of valid responses to a given context. 

Given an objective function, optimization can be done following the \ac{RL} paradigm~\citep{sutton2018reinforcement}, which is successfully applied to physically constrained environments~\citep{silver2016go, levine2016end2end, levine2016learning, finn_corl17}.  The majority of previous work in the area of \iss does this by considering the \is as the agent and the underlying stochastic environment induced by the user~\citep{hofmann2013_wsdm, Li_emnlp_2016, peng2017composite, lipton2018bbq, Su2018D3Q} where the system policies are optimized by interacting with real users or user simulators. However, this setup does not allow us to apply the principle, outlined earlier, that it is a user (not the \is) who is getting reward by interacting with the system while maximizing their utility.
Recently, \citet{leike2018scalable} showed how agent alignment, cast in an \ac{RL} framework, can be applied for optimizing general purpose \iss via reward modeling. \cite{jeon2020reward} and \citet{reddy2019reward} demonstrate how this approach can be applied in the robotics domain. However, this setup requires a quantity of user feedback that may not always be available in practice~\citep{Fox_trans_2005, Joachims_2005}, which leave us with unlabeled user trajectories.

\begin{figure*}[ht!]
\centering
   \includegraphics[clip, width=\textwidth]{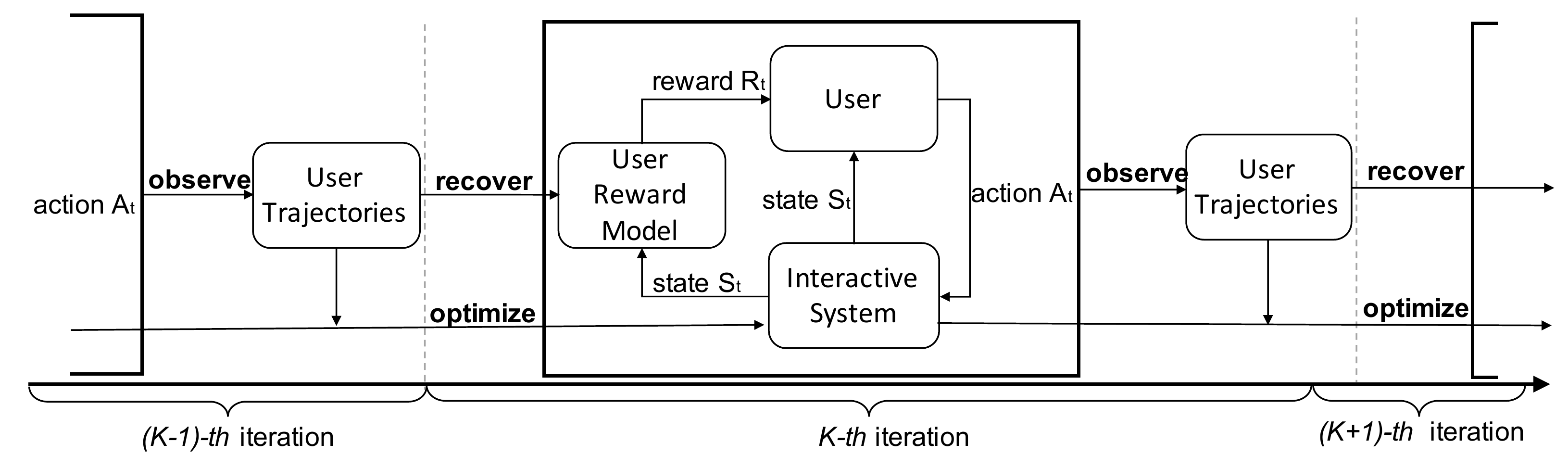}
   \caption{Schematic illustration of the proposed setup of iterative gradual optimization of the flow of the \is: the user reward model is \emph{recovered} from the logs collected while users are interacting with a \is;~\acf{ISO} is used to \emph{optimize} the \is at each iteration.
   }
   \label{fig:iso}
\end{figure*}

In this paper, we assume that users continue their interactions with the system if their goals are fulfilled\footnote{or at least partially fulfilled}, so they are getting rewards after each action.  we propose a general perspective on how to improve \iss by simultaneously \begin{inparaenum}[(1)]\item inferring an objective function directly from data, namely unlabeled trajectories of user interactions with the system, and \item iteratively, and step by step, optimizing the system for this data-driven objective\end{inparaenum}. 
Since users have difficulties in comprehending dramatic changes in an \is \citep{mitchell1989dynamic,white2002finding,obendorf2007web,teevan2008people}, changes should be made gradually so as to let users adapt to a newly optimized interactive system.
The proposed setup is schematically outlined in Figure~\ref{fig:iso}. It embodies a principled approach by concurrently inferring data-driven objectives from user interactions and optimizing the \is accordingly. Thus, our approach does not depend on any domain knowledge.

Below, we start by outlining relevant research areas (Section~\ref{section:relatedwork}).
Then we make the following contributions:
\begin{itemize}[nosep,leftmargin=*]
\item A new way of modeling user-system interactions, which is depicted as $k^\text{th}$ iteration in Figure~\ref{fig:iso} (System~\ref{sec:interactions}).
\item A novel optimization setup to infer data-driven objectives that accurately reflect the users' needs solely from interaction, without using any domain knowledge to handcraft an optimizing goal, which is partially reflected by the arrow `recover' in Figure~\ref{fig:iso} (Section~\ref{sec:objectives}). 
\item A novel algorithm, \ac{ISO}, that optimizes an \is through data-driven objectives, which is depicted as the arrow labeled `optimize' in Figure~\ref{fig:iso} (Section~\ref{sec:opt-method}). 
\item To validate the success of the proposed method, we apply it to two different simulated interactive systems. We show how the proposed optimizer can improve the system performance in the designed setups. We also show that by inferring user reward functions, we can optimize the interactive system without real users in the loop and real users are only involved while collecting user-system interaction trajectories (Section~\ref{section:experimental_all}). 
\end{itemize}

\section{Related Work}
\label{section:relatedwork}

Relevant work for this paper comes in two broad strands: how to optimize interactive systems (Section~\ref{sec:rel_work_optimize}) and what reward signal can be used for optimization (Section~\ref{sec:rel_work_reward}).  

\subsection{Optimizing Interactive Systems}
\label{sec:rel_work_optimize}

The flow of interactive systems~\citep{white2016interactions} can be improved by direct and indirect optimization. Direct optimization aims to maximize user satisfaction directly; in contrast, indirect optimization solves a related problem while hoping that its solution also maximizes user satisfaction~\citep{dehghani2017neural}.
Direct optimization can be performed using supervised learning or \ac{RL}~\citep{mohri2012foundations}. In \ac{RL}, an agent learns to alter its behavior through trial-and-error interactions with its environment~\citep{sutton1998rl}. 
The goal of the agent is to learn a policy that maximizes the expected return. 
\ac{RL} algorithms have successfully been applied to areas ranging from traditional games to robotics~\citep{mnih2015human,silver2016go,levine2016end2end,levine2016learning,duan2016rl,wang2016learning,zhu2017target,schulman2017proximal,haarnoja2018soft,vinyals2019grandmaster,akkaya2019solving,hafner2019dream,schrittwieser2019mastering}.

Many applications of \ac{RL} to optimizing interactive systems come from such fields as \ac{IR}, recommender systems, and dialogue systems. General assumption of users trying to maximize their utility proposed in~\cite{reddy2019reward, jeon2020reward} holds for \iss as well~\cite{azzopardi_sigir_2014}.
\citep{hofmann2011_ecir,hofmann-balancing-2013} apply \ac{RL} to optimize \ac{IR} systems; they use \ac{RL} for online learning to rank and use interleaving to infer user preferences~\citep{hofmann2013_wsdm}. 
\citet{shani2005mdp} describe an early MDP-based recommender system and report on its live deployment.
\citet{Li_emnlp_2016} apply \ac{RL} to optimize dialogue systems; in particular, they optimize handcrafted reward signals such as ease of answering, information flow, and semantic coherence. 
A number of \ac{RL} methods, including Q-learning~\citep{peng2017composite, lipton2018bbq,li2017end,Su2018D3Q} and policy gradient methods \citep{dhingra2016towards,williams2017hybrid,takanobu2019guided}, have been applied to optimize dialogue policies by interacting with real users or user simulators. With the help of \ac{RL}, the dialogue agent is able to explore contexts that may not exist in previously observed data. 
A key component in \ac{RL} is the quality of the reward signal used to update the agent policy. 
Most existing \ac{RL}-based methods require access to a reward signal from user feedback or a predefined reward. 

However it still remains non-trivial to apply \ac{RL} paradigm towards scalable real world machine learning tasks~\citep{leike2018scalable} due to the lack of general approach of recovering data-driven objectives, which we discuss next.

\subsection{Rewards for Interactive Systems}
\label{sec:rel_work_reward}

When applying \ac{RL} to the problem of optimizing interactive systems, we need to have rewards for at least some state-action pairs. 
Previous work typically handcrafts those, using, e.g., normalized discounted cumulative gain (nDCG)~\citep{odijk-dynamic-2015} or clicks~\citep{kutlu2018correlation,zhao2018recommendations} before the optimization or the evaluation of the algorithm.
Instead of handcrafting rewards, we recover them from observed interactions between the user and the interactive system using \ac{IRL}.
The main motivation behind \ac{IRL} is that designing an appropriate reward function for most \ac{RL} problems is non-trivial; this includes animal and human behavior~\citep{abbeel_icml_2004}, where the reward function is generally assumed to be fixed and can only be ascertained through empirical investigation.
Thus inferring the reward function from historical behavior generated by an agent's policy can be an effective approach.
Another motivation comes from imitation learning, where the aim is to teach an agent to behave like an \emph{expert} agent. 
Instead of directly learning the agent's policy, other work first recovers the expert's reward function and then uses it to generate a policy that maximizes the expected accrued reward~\citep{ng_icml_2000}.
Since the inception of \ac{IRL}~\citep{Russell_1998aa}, several IRL algorithms have been proposed, including maximum margin approaches~\citep{abbeel_icml_2004,ratliff2009learning},
and probabilistic approaches such as~\citep{ziebart_aaai_2008} and \citep{boularias2011relative}. In the last few years,  a number of adversarial IRL methods \citep{finn2016connection,guided_cost_learning,ho2016generative,fu2017learning,qureshi2018adversarial,NIPS2019_9002} have been proposed because of its ability to adapt training samples to improve learning efficiency.  One more aspect aspect \ac{IRL} methods are differ is availability of feedback or score for the user trances. \cite{christiano2017deep, leike2018scalable} suggest setup where system can learn from user feedback which is not always available in practice. In our paper, we tackle the case with no explicit feedback.

Regarding of the applications of \ac{IRL}, \citet{ziebart_iui_2012} use \ac{IRL} for predicting the desired target of a partial pointing motion in graphical user interfaces.
\citet{monfort_aaai_2015} use \ac{IRL} to predict human motion when interacting with the environment. 
\ac{IRL} has also been applied to dialogues to extract the reward function and model the user~\citep{pietquin2013inverse,takanobu2019guided,li2020guided,li2019dialogue}. 
\ac{IRL} is used to model user behavior in order to make predictions about it. But
we use \ac{IRL} as a way to recover the rewards from user behavior instead of handcrafting them and optimize an interactive system using these recovered rewards.
\citet{lowe2017towards} learns a function to evaluate dialogue responses. However, the authors stop at evaluation and do not actually optimize the interactive system.

Recent work~\citep{leike2018scalable, zhang2019learning,jeon2020reward} demonstrate impressive results and outline new research direction while modeling user-system interaction using the agent alignment problem~\citep{sutton2018reinforcement}. In contrast to our work, the reward modeling heavily relies on a user feedback loop, which is mostly not available in the internet based \iss~\citep{kiseleva_cikm_2014,kiseleva_serp_sigir_2015}.

\medskip

The key difference between our work and previous studies is that we first use recovered rewards from observed user interactions to reflect user needs and define \is objectives. Subsequently the \is can be optimized according to the defined data-driven objectives so as to improve the user experience.  We regard the interactions between a user and an \is as an agent interacting with an changeable environment, where the transition distribution of the environment can be updated. 
Treating an interactive system as a changeable and programmable environment is novel and reasonable because we have complete control on the behaviors of \iss since we are the system designers. 
\citet{lowe2017towards} learned a function to evaluate dialogue responses but does not actually optimize the interactive system. \citet{leike2018scalable} formulated the optimization problem in a complex multi-agent setup because their environment is physical and non programmable. Besides, the reward modeling by \citep{leike2018scalable} heavily relies on user feedback loop, which is mostly not available in the \iss.

\section{Modeling User-System Interactions}
\label{sec:interactions}

In this section, we first introduce our assumptions about collaborations between a user and an \is (Section~\ref{sec:assumption}), and then we explain how we model these interactions (Section~\ref{sec:mdp}). 

\subsection{Assumptions}
Our goal is to design an \is that can successfully assist users with completing some real world tasks.
We have formulated a set of assumptions to formalize user-system interactions, which are schematically depicted in Figure~\ref{fig:iso}. They can be roughly separated into two groups: assumptions about the system design (S) and assumptions regarding to user goals and behavior (U):
\label{sec:assumption}
\begin{enumerate}[label=\textbf{Assumption \arabic*},ref= Assumption \arabic*,leftmargin=*]
\item \label{ass:system-goal} \textbf{S:} system's goal is to accommodate better user experience, namely maximize utility a user gets from the system by minimizing their efforts;
\item \label{ass:system-setup} \textbf{S:} system setup allows us to iteratively and gradually improve the system in a sequential manner to accommodate better user experience, given that at the beginning the system provides `non-zero' utility for users but can be significantly improved;
\item \label{ass:system-const} \textbf{S:} a system designer has the ability to transform an \is, but it has some obligatory steps a user needs to take to complete their task due to system design constraints;
\item \label{ass:user-insentive} \textbf{U:} users have incentives to continue their iterations with a \is if they are getting some value from it;
\item \label{ass:user-behave} \textbf{U:} users of a \is have \emph{approximately} homogeneous behaviour, namely users have a shared notion of utility that can be approximated by some objective function \footnote{Terms `users' and `a user' are used interchangeably.};
\item \label{ass:user-max} \textbf{U:} users try to maximize their utility while interacting with a system;
\item \label{ass:user-no-feedback} \textbf{U:} users are not required to provide feedback about their experience. However, user actions can be considered as \emph{implicit signals} reflecting their satisfaction/frustration with an \is.

\end{enumerate}

\subsection{Modeling Interactions}
\label{sec:mdp}
While employing \acf{RL} formalism our~\ref{ass:user-behave} and \ref{ass:user-max} can be reformulated as follows: the user is seen the \emph{optimal} agent who interacts with the environment, an \is, with the goal of maximizing their expected rewards.

As a \emph{running example} we can consider a user who is interacting with a search engine.
The process of user-system interaction is modeled using a finite \ac{MDP} $(S, A, T, r, \gamma)$, in the following way:\footnote{We follow the notation proposed in~\citep{sutton2018reinforcement}}
\begin{itemize}[leftmargin=*]

\item $S$ is a set of states that represent responses from the \is to the user. $S$ is finite as there are limited predefined number of responses that \is can return.
\item $A$ is a finite set of actions that the user can perform on the system to move between states. In case of search engine, a user can run a query, click on the returned results, reformulate a query etc.
\item $T$ is a transition distribution and $T(s, a, s')$ is the probability of transitioning from state $s$ to state $s'$ under action $a$ at time $t$: 
\begin{equation}
    T (s' \mid s, a) =  \mathbb P(S_{t+1}=s' \mid S_t = s, A_t=a).
\end{equation}
For search engines, being at the start page (which is $s$) a user is making an action $a$, e.g.\ running a query, and the engine redirects him to a result page (which is $s'$).

\item $r(s,a,s')$  is the expected immediate reward after transitioning from $s$ to $s'$ by taking action $a$. 

We compute the expected rewards for (state, action, next state) triples as:
\begin{equation}
\mbox{}\!\!
r(s,a,s') =  \mathbb{E}[R_t \mid S_t = s, A_t=a,  S_{t+1}=s'],    
\end{equation}
where $R_t$ is reward at time $t$. In case of search engine, a user is getting a reward for finding a desired information. However, the rewards are not observed in practise (\ref{ass:user-no-feedback}).
For simplicity in exposition, we write rewards as $r(s)$ rather than $r(s, a, s')$ in our setting; the extension is trivial~\citep{ng_icml_2000}.
\item $\gamma \in (0,1]$ is a discount factor.
\end{itemize}

We write $\mathcal{P}$ to denote the set of interactive systems, i.e., triples of the form $(S,A,T)$. 
Following \ref{ass:system-const}, system designers have control over the sets $S$, $A$, and the transition distribution, $T$, and $T$ can be changed to optimize an \is.

The \emph{user behavior strategy} for accomplishing their tasks is represented by a policy, which is a mapping, $\pi \in \Pi$, from states, $s \in S$, and actions, $a \in A$, to $\pi(a|s)$, which is the probability of performing action $A_t=a$ by the user when in state $S_t=s$. The observed history of interactions between the user and the \is, $H$\footnote{$H$ can be referred further as logs of user interactions, or log, or user trajectories/traces.}, is represented as a set of trajectories, $\{\zeta_i\}_{i=1}^n$, drawn from a distribution $Z$, which is brought about by $T$, $\pi$, and $D_0$, where $D_0$ is the initial distribution of states. following \ref{ass:user-behave}, which proposes homogeneity in user behavior, simplifies the problem, i.e. as if one user generated $H$. 
A \emph{trajectory} is a sequence of state-action pairs, where a user does not provide explicit feedback (\ref{ass:user-no-feedback}):
\begin{equation}
\label{eq:traj}
\zeta_i = S_0,A_0, S_1,A_1, \dots, S_t, A_t, \dots. 
\end{equation}

To conclude, we suppose that the user is an optimal agent who is trying to maximize its reward 
under the system dynamics it faces and that the system wants to improve the user experience over time 
by creating progressively easier MDPs to solve for the user. However, a \is cannot transition from all initial to goal states in one step due to design constraints. For example, if a user is searching for holiday destinations, the system cannot redirect him to the final stage of booking a hotel because he needs to go through a necessary step, e.g.,\ providing payment details.

To summarize, we have described the basic principles of modeling interactions between users and an \is.
Next, we detail how to define data-driven objectives that are used to optimize an \is.

\section{Defining Data-driven Objectives}
\label{sec:objectives}
In this section, we first present our approach to convert user needs to data-driven objectives of an \is (Section~\ref{sec:def_goals}), and then we explain how these objectives can be estimated (Section~\ref{sec:compute_goals}).
\subsection{Defining Interactive System Objectives}
\label{sec:def_goals}

We define the \emph{quality} of an \is as the expected state value under an optimal user policy. The value of a state $S_0$ under a policy $\pi$ is given as~\citep{sutton2018reinforcement}:
\begin{equation}
  V^{\pi} (S_0) =  \mathbb{E}_{\pi} \left[\sum^{\infty}_{t=0} \gamma^t R_{t+1}\right],
\end{equation} 
where the expectation $\mathbb{E}_{\pi}[\cdot]$ is taken with respect to sequences of states $S_0, S_1, \dots, S_t, \dots$ drawn from the policy $\pi$ and transition distribution $T$. 
We use $V^\pi_T$ to denote the value of a policy $\pi$ under the current transition distribution $T$, and hide the initial states $S_0$ for simplification.

In the proposed setting, the user goal is to find the best policy $\pi^*$ such that  $V^\pi_T$  is maximized.
 $V_*(T)$ defines the maximum possible value of $V^\pi_T$ under transition distribution T as follows:
\begin{equation}
\label{eq:pi_value_best}
V_*(T)=\max_{\pi \in \Pi} \limits V^\pi_T, 
\end{equation}
where $\Pi$ is the set of possible user policies.
We formulate the problem of finding the optimal \is's transition distribution, denoted $T^*$, in the following terms:
\begin{equation}
\label{eq:optimal_environment}
T^* = \argmax _{T \in T}V_*(T).
\end{equation}
Therefore, \optimize represents the objective function, mentioned in \ref{ass:user-max}, which is derived from user trajectories (Eq.~\ref{eq:traj}) directly. After finding $T^*$ through solving the proposed optimization problem, the system designer has the ability to transform the current system to a new one, which should deliver a better user experience as it reflects user needs better. This process is illustrated in Figure~\ref{fig:iso} by the arrow marked \emph{optimize} between two consecutive iterations.

With the transition distribution $T$, the interactive system will respond with the next state $s'$ given the current state $s$ and the user action $a$. In real life, it is not guaranteed that the tuple $(s,a,s')$ exists. For example, in task-oriented dialog systems, the  system first needs to collect essential information for booking a hotel (e.g., hotel name, room type) step by step. In some cases, the system also needs to recommend potential hotels and asks the user to make a choice. After successfully collected all information, the system can guide the user to a payment page. Obviously, it is not possible to deliver a payment state to the user when the information contained in the current state is not complete. Therefore, inherent constraints exist in interactive systems and this makes finding the optimal interactive system's transition distribution a meaningful and interesting task. Otherwise, the system can always deliver the most valuable state to the user in one step at any states.        

To estimate the data-driven objectives \is presented in~\optimize, we first need to recover $R_t$, which we will discuss next. 

\subsection{Recovering User Rewards}
\label{sec:compute_goals}

\ref{ass:user-insentive} suggests that continued user interactions with the system indicate a certain level of user satisfaction, which can be reflected by experienced rewards.
In contrast with $\zeta_i \in H$ presented in Eq.~\ref{eq:traj}, the complete history of interactions, $\hat{H}$, consists of trajectories $\hat{\zeta_i} \sim \hat{Z}$, which include the user reward $R_t$:
\begin{equation}
\label{eq:scored_trajectory}
\hat{\zeta_i} = S_0,A_0,R_1, S_1,A_1,R_2 \dots, R_t, S_t, A_t, \dots.
\end{equation}

The problem is that the true user reward function is hidden from a \is and inherently difficult due to the complexity of the real world surrounding users. 
Our goal is using the collected incomplete user trajectories, $H$, shown in Eq.~\ref{eq:traj}, to find a way to approximate true user rewards. 
To address this challenge we apply \acf{IRL} methods\footnote{\ac{IRL} methods are described in greater details in Section~\ref{sec:rel_work_reward}}, which are proposed to recover the rewards of different states, $r(s)$, for $\zeta_i \in H$.
Our assumption about the form of user reward function is:
given state feature functions $\phi : {S_t} \to \mathbb{R}^k$ that describe $S_t$ as a $k$-dimensional feature vector, the true reward function $r(s)$ is a linear combination of the state features $\phi(s)$, which can be given as $r(s) = \theta^T \phi(s)$. To uncover the reward weights $\theta$, we employ the following approaches. \\

\noindent
\textbf{\ac{MaxEnt-IRL}:}
The core idea of\\ \ac{MaxEnt-IRL}~\citet{ziebart_aaai_2008} is that trajectories with equivalent rewards have equal probability to be selected and trajectories with higher rewards are exponentially more preferred, which can be formulated as: 
\begin{eqnarray}
\mbox{}\hspace*{-1mm}&&\mbox{}\hspace*{-1mm}
\smash{\mathbb{P}(\zeta_i\mid \theta) = \frac{1}{Z(\theta)} \exp(\theta^T\phi({\zeta_i})) = \frac{1}{Z(\theta)} \exp(\sum_{t=0}^{|\zeta_i|-1}\theta^T\phi(S_t)),}
\label{eq:maxent-irlj}\\
\nonumber
\end{eqnarray}

\noindent
where $Z(\theta)$ is the partition function. 
\ac{MaxEnt-IRL} maximizes the likelihood of the observed data under the maximum entropy (exponential family) distribution. \\

\noindent
\textbf{\textbf{\ac{AIRL}:}}
Based on \ac{MaxEnt-IRL}, \citep{guided_cost_learning} combine sample-based \ac{MaxEnt-IRL} with forward reinforcement learning to estimate the partition function $Z$, where: 

\begin{eqnarray}
\mbox{}\hspace*{-1mm}&&\mbox{}\hspace*{-1mm}
L(\theta) = 
\smash{-\E_{\zeta_i \sim p} r_\theta(\zeta_i) + \log \left(\E_{\zeta_j \sim q} \left[\frac{\exp(r_\theta(\zeta_j))}{q(\zeta_j)}\right]\right).}
\label{eq:guided-cost-obj}\\
\nonumber
\end{eqnarray}

Here, $r_\theta(\zeta_i)$ is the reward of trajectory $\zeta_i$, $p$ represents the distribution of demonstrated samples, while $q$ is the background distribution for estimating the partition function $\int \exp(r_\theta (\zeta)) d \zeta$. 
Due to high variance from operating over entire trajectories, \citet{fu2017learning} extend the algorithm to single state-action pairs and the proposed method, \ac{AIRL}, which is a practical and scalable \ac{IRL} algorithm based on an adversarial reward learning formulation. We use \ac{AIRL} to recover the reward function for complex interactive systems since \ac{AIRL} can estimate non-linear reward functions. \\

\noindent
\textbf{\ac{DM-IRL}:}
For completeness, we also employ \ac{DM-IRL}~\citep{el2013_dm_irl, Burchfiel_aaai_2016}, which deal with scored trajectories, to have a case of perfectly recovered reward weight $\theta$ for comparison. \ac{DM-IRL} directly attempts to regress the user's actual reward function that explains the given score.
\ac{DM-IRL} uses discounted accrued features to represent the trajectory:
$\psi(\zeta_i) = \sum_{t=0}^{|\zeta_i|-1} \gamma^t \phi(S_t)$, where $\gamma$ is the discount factor. The score of a trajectory $\zeta_i$ is $\text{score}_{\zeta_i} = \theta^T \psi(\zeta_i)$.
Since the exact score for each trajectory is supplied, the recovered rewards with \ac{DM-IRL} are exactly the ground truth of reward functions, which can be regarded as oracle rewards.\\
This process is depicted in Figure~\ref{fig:iso} by the arrow marked \emph{recover}. Once we have recovered the reward function $r(s)$, we can proceed to the optimization objectives presented in~\optimize.

\section{Optimizing Interactive Systems}
\label{sec:opt-method}

We start by explaining how to maximize the quality of an \is for a user behaving according to a fixed stationary policy $\pi$:
\begin{equation}
\label{eq:optimal_environment_fixed_policy}
T^*_\pi = \argmax _{T \in T}V^\pi_T.
\end{equation}

To solve this problem, we first build an MDP as proposed above, where the user is the agent and the system is the environment Following~\ref{ass:system-setup}, the system can be optimized to improve the user experience which we characterized by the quality of the \is.
This problem is \emph{equivalent} to finding the optimal policy in a reformulated $\text{MDP}^+(S^+,A^+, T^+,  r^+,\gamma^+)$, where the agent is an \is and the stochastic environment is a user. It should be noted that the roles of agent and environment in the reformulated $\text{MDP}^+$ are exactly the opposite of the roles in the original MDP. We also convert the state space and action space correspondingly. We rely on the first MDP for inferring the user reward functions, while we rely on the second one, $\text{MDP}^+$, for updating \iss.
In $\text{MDP}^+$, the state $S_t^+ $ is represented by a concatenation of the state $S_t$ the user is in and the action $A_t$ the user takes at time step $t$ from the original MDP; the action $A^+_t$ is the original state $S_{t+1}$.
The \is observes the current state $S^+_t$ and picks an action $A^+_t$ under the \is policy $\pi^+(A^+_t|  S^+_t)$. Then the user returns the next state $S_{t+1}^{+}$ according to the transition distribution $T^+ (S_{t+1}^+ | S_t^+, A_t^+)$ which is inferred from the policy model $\pi (A_{t+1}  |  S_{t+1})$. 
Therefore, finding the optimal transition $T^{*}_{\pi}$ from Eq.~\ref{eq:optimal_environment_fixed_policy} is equivalent to finding the optimal policy $\pi_{*}^+$ in the reformulated $\text{MDP}^+$ as follows:

\begin{equation}
\label{eq:optimal_policy_plus}
  \pi^{+}_{*} = \argmax _{\pi^+ \in \Pi^+} V^{\pi^+}_{T^+} ,
  \end{equation}

which can be done using an appropriate \ac{RL} method such as Q-learning or Policy Gradient. $D_0^+$ is the initial distribution of states in $\text{MDP}^+$.
After we have demonstrated how to optimize the \is for a given stationary policy, we return to the original problem of optimizing the \is for an optimal policy $\pi_*$. 

\begin{algorithm}
   \caption{Interactive System Optimizer (ISO)}
   \label{alg:opt-env}
   \begin{algorithmic}[1]
   \STATE {\bf Input:} Original system $(S, A, T)$, $r$, $\gamma$, $D_0$.
   \STATE  Construct original $\text{MDP}$$(S, A, T, r, \gamma)$ 
   \STATE  $\pi_*(a|s) = RL (S, A, T, r, \gamma)$ \textit{    // finding the current user policy}
   \STATE Construct system MDP$^+(S^+, A^+, T^+, r^+, \gamma^+)$: \textit{// reformulate the original MDP by switching the roles of agent and environment}
               \begin{itemize}[nosep]
               \item $S_t^+ = S_t \oplus A_t$ \textit{// build the new state space by concatenation}
               \item $A_t^+ = S_{t+1}$ \textit{// build the new action space}
               \item $T^+ (S_{t+1}^+ | S_t^+, A_t^+)  = 
                          \pi_*(A_{t+1} | S_{t+1})$  \textit{// build the transitions in $MDP^+$}
               \item $r(S_t^+)^+$ = $r(S_t)$ \textit{// convert the reward function}
               \item $\gamma^+$ = $\gamma$ \textit{// both MDPs share the same discount factor}
             \end{itemize}
   \STATE $D_0^+ \sim ( S_0 \sim D_0 , A_0 \sim \pi_*(a| S_0))$ \textit{// sample initial states in$MDP^+$}
   \STATE $\pi^+(A_t^+|S_t^+) = T(S_{t+1}|S_t, A_t)$   \textit{// find the optimal transition distribution in the original MDP is formulated as fining the optimal policy in a reformulated $MDP^+$}      
   \STATE $\pi^{+}_*(a^+|s^+)  = RL (S^+, A^+, T^+, r^+, \gamma^+)$  \textit{// optimize the system policy in $MDP^+$}        
   \STATE $T^*(S_{t+1} | S_t, A_t)  = \pi^{+}_* (A_t^+ | S_t^+) $ \textit{// replace the transition distribution in original MDP with the newly updated system policy}
   \STATE {\bfseries Output:} Optimized system $(S, A, T^*)$ 
\end{algorithmic}
\end{algorithm}

To summarize, we propose a formal procedure for optimizing \iss, called \ac{ISO}, presented in Algorithm~\ref{alg:opt-env}, with the following steps:
\begin{enumerate} [label=\textbf{Line \arabic*},ref= Line \arabic*,leftmargin=*]
    \item We assume that we have an estimate of the reward function $r(s)$ using one of the \ac{IRL} methods described in Section~\ref{sec:compute_goals}. So we have as input: the original system $(S, A, T)$, the reward function $r$, the discount factor $\gamma$, and the initial distribution of states $D_0$.
    \item \ac{ISO} formulates the original system as $\text{MDP}(S, A, T, r, \gamma)$.
    \item  \ac{ISO} uses an appropriate \ac{RL} algorithm to find the current user policy $\pi_*(a|s)$ given the reward function $r$.   
    \item  \ac{ISO} transforms the original MDP$(S, A, T, r, \gamma)$ into the new one\\ $\text{MDP}^+(S^+,A^+, T^+ , r^+, \gamma^+)$, where the roles of the agent and environment are switched. In our setting, $S^+_t$ has the same reward value as $S_t$. 
    The discount factor $\gamma^+$ remains the same.
    \item \ac{ISO} transforms $D_0$ to $D_0^+$ to match the distribution of first state-action pairs 
    \item The equivalence $\pi^+(A_t^+ |  S_t^+) = T(S_{t+1} |  A_t,S_t)$ means that finding the optimal $\pi^+_{*}$ according to Eq.~\ref{eq:optimal_policy_plus} is equivalent to finding the optimal $T^*_{\pi}$ according to Eq.~\ref{eq:optimal_environment_fixed_policy}. Therefore, the transition distribution can be regarded as a policy network or a policy table from $\text{MDP}$'s perspective depending on the policy learning method.
    \item We can use an appropriate \ac{RL} algorithm to find $\pi_*^+(A_t^+ |  S_t^+)$.
    \item  \ac{ISO} extracts $T^*(S_{t+1} |  S_t, A_t)$ from the optimal system policy $\pi_*^+(A_t^+ |  S_t^+)$. The extraction process is trivial: $T^*(S_{t+1} |  S_t, A_t) =  \pi^+(A_t^+ |  S_t^+)$. Then, \ac{ISO} terminates by returning the \emph{optimized} \is.
    \item  \ac{ISO} outputs the optimized \is $(S, A, T^*)$.
\end{enumerate}

Once \ac{ISO} has delivered the optimized system $(S, A, T^{*})$, we expose it to users so they can interact with it as illustrated in Figure~\ref{fig:iso}. 
We assume that users adjust their policy to $T^{*}$. 
After enough iterations the user policy will converge to the optimal one.
Iterations between optimizing the \is for the current policy and updating the user policy for the current \is continue until both converge.

In summary, we have presented the \acf{ISO}. It optimizes an \is using data-driven objectives. It works by transforming the original \ac{MDP}, solving it and using its solution to yield the optimal transition distribution in the original \ac{MDP}. 

\section{Experiments and Results}
\label{section:experimental_all}
In this section, we apply our proposed method, \ac{ISO}, to two different simulated interactive setups. In the first setup, the \is operates in a tabular-based world with finite states and actions (Section~\ref{sec:exp-tabluar}). The second one has a more realistic setup, where the agent, the environment, and the reward function are all represented by neural separate networks (Section~\ref{sec:exp-neural}). Each proposed experimental setup is described with regard to the three following components: the design of the \is, modeling user behavior, and suitable evaluation process.
For both experimental setups, our results demonstrate that \ac{ISO} can improve significantly improve the system performance in the designed setups. We conclude this section openly discussing a list of foreseeing limitations (Section~\ref{sec:limitations}).

\subsection{Optimizing Interactive Systems in a Tabular-based World}
\label{sec:exp-tabluar}
\subsubsection{Experimental Setup}
\label{section:experimentalsetup}

\paragraph{\textbf{Designing an Interactive System}}
We simulate an arbitrary \is where we need a finite set of states $S$, a finite set of actions $A$ and a transition distribution $T$. 
Features of a state $\phi(s)$ are fixed. 
For our experimental setup, we simulate the \is where $|S| = 64$ and $|A| = 4$. 
We work with a complex environment where a user can transition between any two states if these two states are connected. 
The connections between two states are predefined and fixed, but the transition distribution is changeable. In another word, for the same system in different runs, the connectivity graph of this systems is fixed and will not be changed once it is sampled at the very beginning. This setup corresponds to the inherent constraints between state transitions in real \iss Section  \ref{sec:def_goals}.
We use a hyper-parameter, the connection factor $cf$, to define the number of possible next states after the user has taken one specific action at the current state.
For an initial \is, $D_0$ is randomly sampled as well as $T$. At each iteration \ac{ISO} delivers $T^*$, which substitutes the initial $T$ obtained at the previous iteration. 
The \emph{optimized} \is is used for the next iteration until the process converges.

\paragraph{\textbf{Modeling User Behavior}}
To model user behavior we require a true reward function $r_{real}(s)$, and an optimal user policy $\pi^{*}_{user}$. 
We utilize a linear reward function $r_{real}(s)$ by randomly assigning $25\%$ of the states reward 1, while all others receive 0. As we use one-hot features for each state, $r_{real}(s)$ is guaranteed to be linear. 

We use a soft value iteration method~\citep{ziebart2010modeling} to obtain the optimal user policy $\pi^*_{user}$.

The quality of the recovered reward functions is influenced by how trajectories are created, which in turn can affect the performance of \ac{ISO} as it relies on $r_{real}(s)$ to optimize the transition distribution $T$ behind the \is with reinforcement learning.  

We experiment with the following types of user trajectories:

\begin{itemize}[leftmargin=*]
\item \emph{Optimal:} Users know how to behave optimally in an interactive system to satisfy their needs. To simulate the user interactions $H$, we use $\pi^{*}_{user}$ trained with the real reward function $r_{real}(s)$.
\item \emph{SubOptimal:} Not all users know the system well, which means that the demonstrated behavior is a mixture of optimal and random. We propose two different methods to simulate suboptimal behavior. The degree of optimality of user behavior is controlled by either of two following factors: (1) the proportion of random behavior (this is called `wandering' behavior in \citep{white2005evaluating}); or
(2) the user action noise, which are collectively called the noise factor (NF) $\in [0.0, 1.0]$.
\begin{description}
\item[\emph{Mix of Behaviors (MB):}] The log of user interactions $H$ is a mix of trajectories generated by the optimal policy and the adversarial policy\footnote{To model suboptimal user behavior we use two user policies: (1)~an optimal user policy $\pi^{*}_{user}$; and (2)~an adversarial policy ($1-\pi^{*}_{user}$), which means we choose the action that has the lowest likelihood according to $\pi^{*}_{user}$. We include an adversarial policy instead of a random one because it is the hardest case as users behave opposite of what we expect. E.g., $\mathit{NF} = 0.2$ means that 20\% of the trajectories are generated with the adversarial policy.}.
\item[\emph{Noise in Behavior (NB):}] In this case, the trajectories in $H$ are generated from the optimal policy but we add noise to the user actions to get suboptimal behavior \footnote{E.g., $\mathit{NF} = 0.2$ means the probability is 20\% that the user will not choose the action with the highest probability in the optimal policy.}. 
\end{description}
\end{itemize}

The generated history of user interactions $H$ represents the case of trajectories without a score which will be fed to \ac{MaxEnt-IRL}.
In terms of \ac{DM-IRL}, interaction history should be given along with scores for each trajectory --  $\hat{H}$. To generate the required dataset $\hat{H}$, we calculate the score using the true reward function $r_{real}(s)$. $\hat{H}$ is the input to \ac{DM-IRL}.

At each iteration, we sample the following datasets reflecting different types of history of user interactions: $\hat{H}$, $H_{Optimal}$, $H_{SubOptimal-0.2-MB}$,  $H_{SubOptimal-0.6-MB}$, $H_{SubOptimal-0.2-NB}$, $H_{SubOptimal-0.6-NB}$  each of size $15,000$ and $|\zeta_i|$ $\in [30, 40]$.

\paragraph{\textbf{Evaluation Process}}
To evaluate the performance of \ac{ISO}, we report the \textit{expected state value} under optimal policy Eq. \ref{eq:pi_value_best} for an \emph{initial} \is (S, A, $T_{init}$) and an \emph{optimized} one (S, A, $T_{opt}$), which we derive after around 100 \textit{iterations} (one \textit{iteration} means we sequentially recover the reward function and run Algorithm \ref{alg:opt-env} once). 

A higher expected state value means users are more satisfied while interacting with the \is. We initialize $40$ different initial \iss by randomly sampling reward functions and transition distribution, and report the overall performance over these $40$ systems.

The true reward functions and the connectivity graphs of these sampled systems are fixed in the whole optimizing process. We use the recovered reward function with \ac{DM-IRL} as the oracle reward for in this setup.

\subsubsection{Results and Discussion}
\label{sec:results}

\begin{figure*}[ht]
\centering
\vspace{-2mm}
   \includegraphics[clip, width=0.8\textwidth]{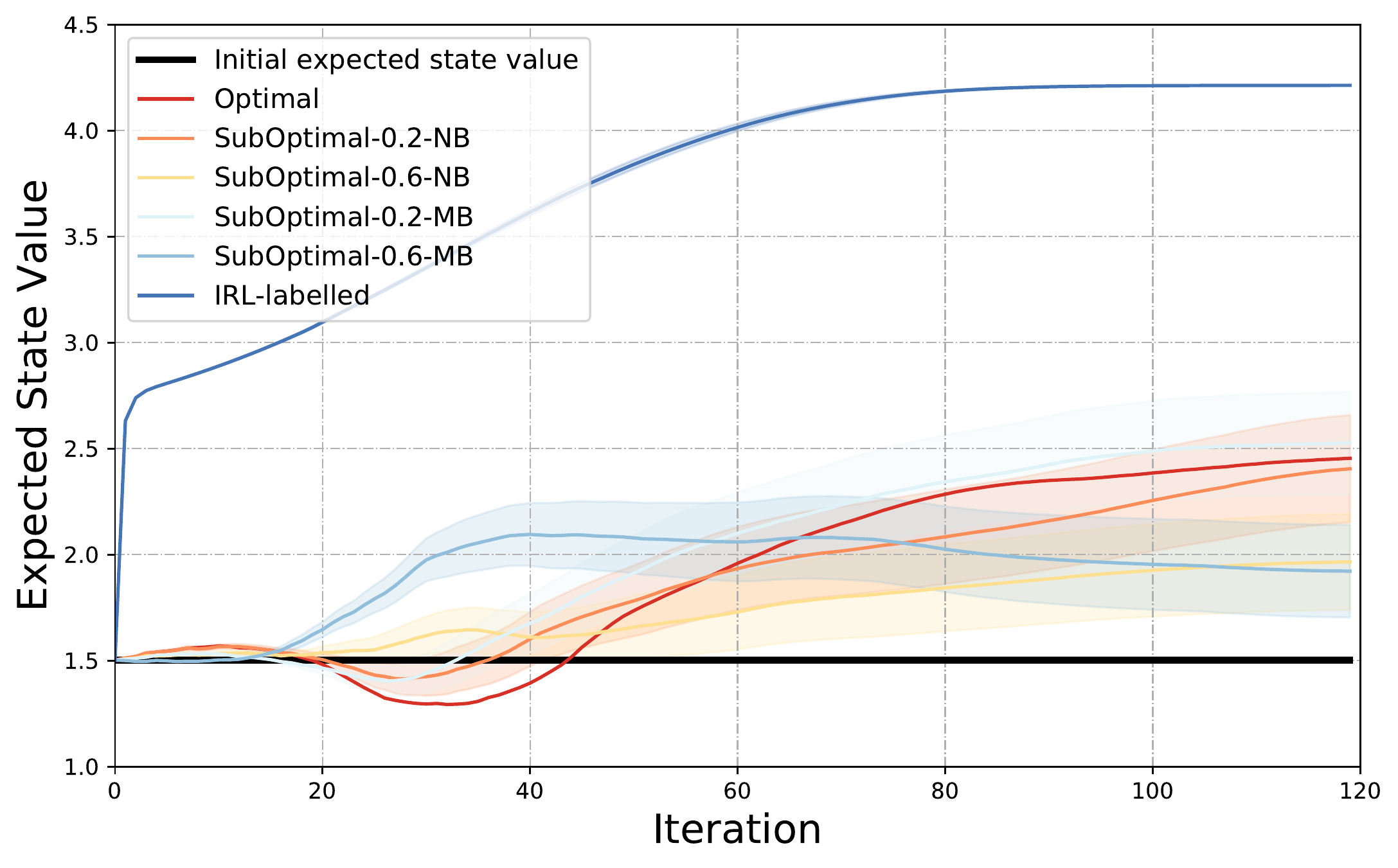}
   \caption{Performance of ISO over 40 randomly sampled systems when connection\_factor=32. The error bounds denote the standard error of the mean ($\pm$SEM). The x-axis is the number of iterations of ISO and the y-axis is the expected state value.}
   \label{fig:state_value_0}
   \vspace*{-0.5\baselineskip}
\vspace{-2mm}   
\end{figure*}

\begin{figure*}[ht]
\centering
\vspace{-2mm}
   \includegraphics[clip, width=0.8\textwidth]{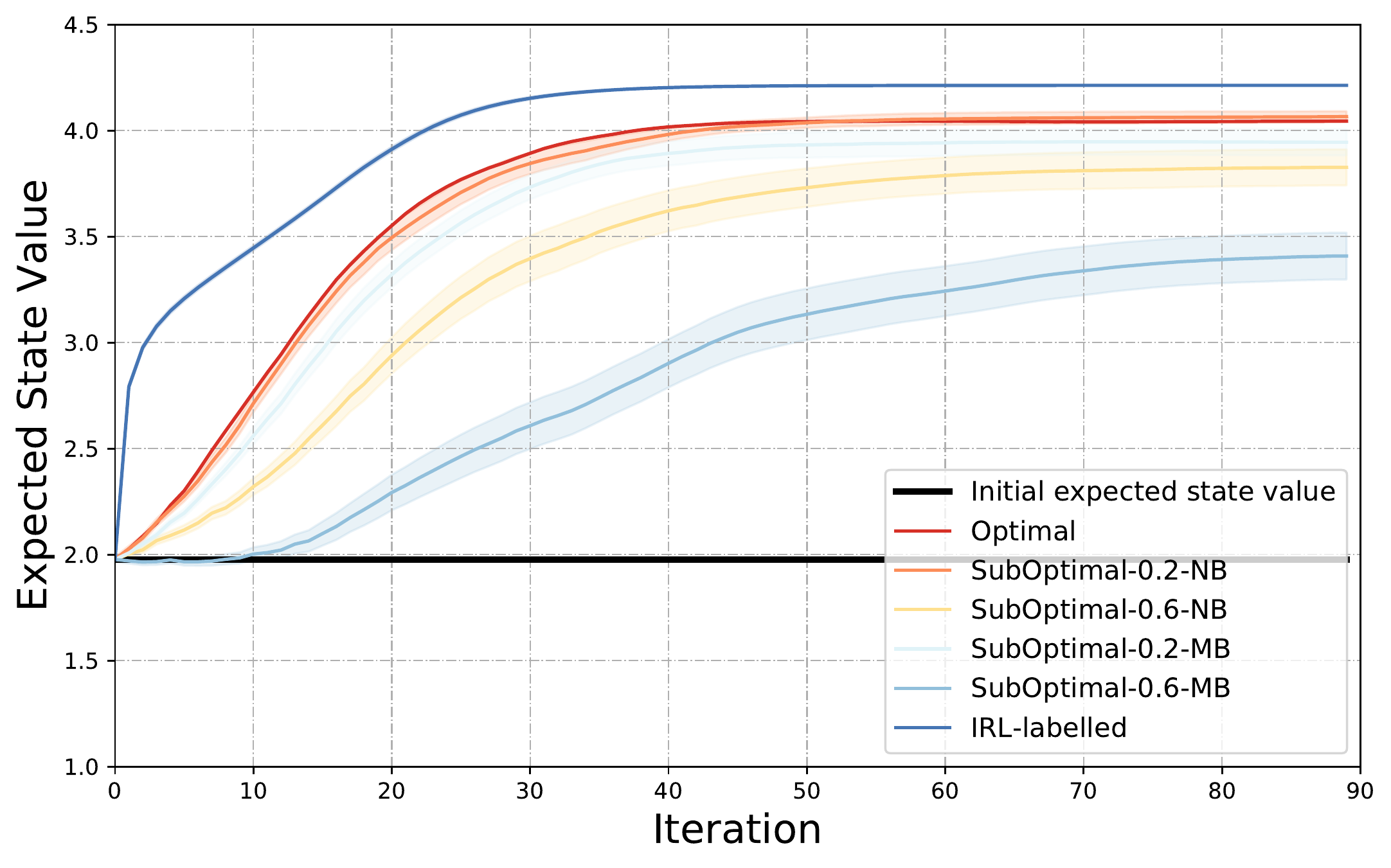}
   \caption{Performance of ISO over 40 randomly sampled systems when connection\_factor=8. The error bounds denote the standard error of the mean ($\pm$SEM). The x-axis is the number of iterations of ISO and the y-axis is the expected state value.}
   \label{fig:state_value_1}
   \vspace*{-0.5\baselineskip}
\vspace{-2mm}   
\end{figure*}

\begin{figure*}[ht]
\centering
\vspace{-2mm}
   \includegraphics[clip, width=0.8\textwidth]{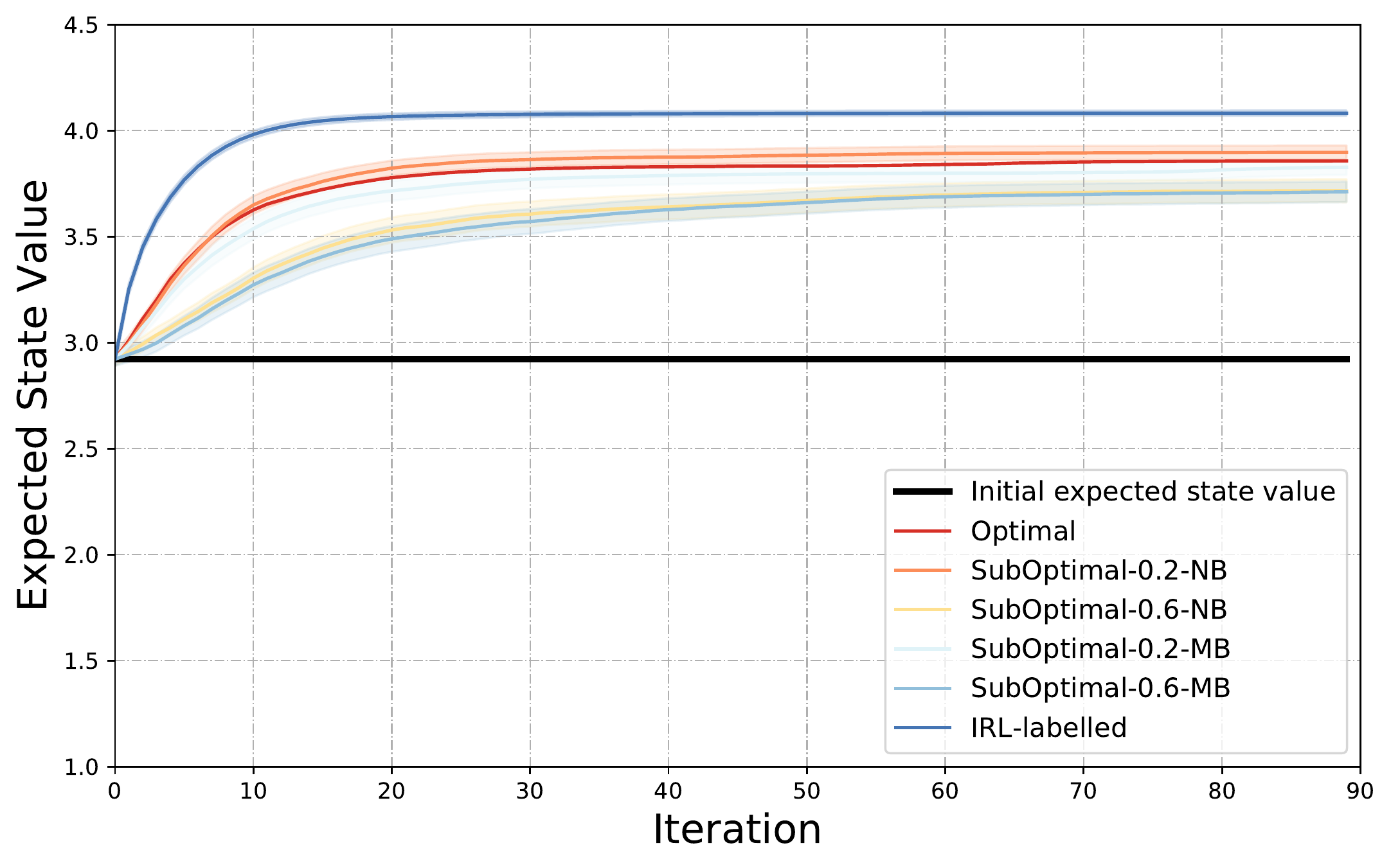}
   \caption{Performance of ISO over 40 randomly sampled systems when connection\_factor=2. The error bounds denote the standard error of the mean ($\pm$SEM). The x-axis is the number of iterations of ISO and the y-axis is the expected state value.}
   \label{fig:state_value_2}
   \vspace*{-0.5\baselineskip}
\vspace{-2mm}   
\end{figure*}

\noindent
\paragraph{\textbf{Improving Interactive Systems with ISO}}
Figure~\ref{fig:state_value_0},\ref{fig:state_value_1},\ref{fig:state_value_2} show how the quality of the interactive system increases with each iteration of \ac{ISO} in terms of different connection factors. The final relative improvements after optimization can be found in Table \ref{Table:results}. We use IRL-lableled to reprsent the system optimized with the recovered reward function by method \ac{DM-IRL}. As expected, when the user gives feedback about the quality of the trajectories (IRL-labelled), the task is simpler and \ac{ISO} manages to get high improvements with the oracle rewards. 
However, the picture changes when we hide the scores from the trajectories.  
Without scores, \ac{ISO} relies on the optimality of user behavior to recover the reward function. 
As the optimality decreases, so does the behavior of \ac{ISO}, and the performance decays.
With the oracle rewards from \ac{DM-IRL}, \ac{ISO} converges quite fast -- as we can see in Figure~\ref{fig:state_value_1} and Figure~\ref{fig:state_value_2}, after 20 iterations the expected state value begins to plateau. Most improvements happen in the first several iterations. 
Thus, \ac{ISO} works with accurately labeled trajectories, but usually obtaining high-quality scores is intractable and expensive in a real \is because the real rewards are invisible. We report it as the oracle performance in our experiment.

With respect to trajectories without scores, ISO is able to improve the initial expected state value. 

In Figure~\ref{fig:state_value_0}, the influence of the noise factor and types of trajectories (MB or NB) is clear. However, in Figure~\ref{fig:state_value_2} where there are fewer connections between two states, only the convergence speeds of different curves are different but they all converge to the same state value eventually. \ac{ISO} manages to optimize the \is even though the user trajectories are quite noisy. 

More remarkable, the convergence speed and final converged values are different depending on the connection factors. As we can see, it is more difficult to get high performance when there are more connections between different states in the predefined systems. More connections mean that more possible trajectories could be taken and it is intractable for \ac{MaxEnt-IRL} to learn a reward function from this kind of situation. By contrast, in Figure~\ref{fig:state_value_2}, each state-action pair can only have two possible next states and the final average state value is much higher than the system in Figure~\ref{fig:state_value_0}. 


\begin{table*}
  \centering
  \resizebox{\linewidth}{!}{
  \begin{tabular}{ cc*{10}{c}}
    \toprule
  \multirow{2}{*}{\diagbox{CF}{\makecell{BT}} }  & \multicolumn{3}{c}{(a) IRL-labelled} & \multicolumn{3}{c}{(b) Optimal (NF=0.0)} & \multicolumn{3}{c}{(c) SubOptimal-0.2-MB (NF=0.2)} \\
  \cmidrule(r){2-4}
  \cmidrule(r){5-7}
  \cmidrule(r){8-10}
&\makecell{Initial} & \makecell{Optimized} & \makecell{Impr.} &\makecell{Initial} & \makecell{Optimized} & \makecell{Impr.}  &\makecell{Initial} & \makecell{Optimized} & \makecell{Impr.}  \\
\midrule
\makecell{32} & 1.50  & 4.21  & $281\%^*$        & 1.50  & 2.45  & $164\%^*$   &1.50  & 2.52   & $169\%^*$     \\
\makecell{8} & 1.98  & 4.21  & $213\%^*$        & 1.98  & 4.04  & $205\%^*$     & 1.98   & 3.94   & $200\%^*$     \\
\makecell{2} & 2.92  & 4.08  & $140\%^*$        & 2.92  & 3.86  & $132\%^*$     & 2.92   & 3.83   & $131\%^*$     \\
    \bottomrule
  \end{tabular}
  }
\resizebox{\linewidth}{!}{
    \begin{tabular}{ cc*{10}{c}}
    \toprule
  \multirow{2}{*}{\diagbox{CF}{\makecell{BT}} }  & \multicolumn{3}{c}{(d) SubOptimal-0.6-MB (NF=0.6)} & \multicolumn{3}{c}{(e) SubOptimal-0.2-NB (NF=0.2)} & \multicolumn{3}{c}{(f) SubOptimal-0.6-NB (NF=0.6)} \\
  \cmidrule(r){2-4}
  \cmidrule(r){5-7}
  \cmidrule(r){8-10}
&\makecell{Initial} & \makecell{Optimized} & \makecell{Impr.} &\makecell{Initial} & \makecell{Optimized} & \makecell{Impr.}  &\makecell{Initial} & \makecell{Optimized} & \makecell{Impr.}  \\
\midrule
\makecell{32} & 1.50  & 1.92  & $128\%^*$        & 1.50  & 2.40  & $160\%^*$   &1.50  & 1.96   & $131\%^*$     \\
\makecell{8} & 1.98  & 3.41  & $173\%^*$        & 1.98  & 4.06  & $206\%^*$     & 1.98   & 3.83  & $194\%^*$     \\
\makecell{2} & 2.92  & 3.71  & $127\%^*$        & 2.92  & 3.90  & $134\%^*$     & 2.92   & 3.72   & $128\%^*$     \\
    \bottomrule
  \end{tabular}
  
  }
\caption{The performance of \ac{ISO}, measured as relative improvement (Impr.) in  expected state value over the Initial \is of the Optimized version (after 120 and 90 iterations) for different types of user behaviors: (a)~IRL-labelled, (b)~Optimal, (c)~SubOptimal-0.2-MB, (d)~SubOptimal-0.6-MB, (e)~SubOptimal-0.2-NB, (f)~SubOptimal-0.6-NB. Only IRL-labelled has access to the trajectory labels. * indicates statistically significant changes ($p<0.01$) using a paired t-test over the initial expected state value and the optimized expected state value.} 
\label{Table:results}

\end{table*}

\paragraph{\textbf{Impact of \ac{ISO} Components}}
The performance of \ac{ISO} depends on its two components: (1)~\ac{RL} methods used to optimize the user policy $\pi_{user}$ for the original \ac{MDP} and system policy $\pi^+_{sys}$ for the reformulated MDP$^{+}$; and (2)~\ac{IRL} methods -- to estimate the true reward function $r_{real}(s)$.
The dependence on \ac{RL} methods is obvious -- the end result will only be as good as the quality of the final optimization, so an appropriate method should be used.
The performance of \ac{ISO} can be influenced by the quality of the recovered reward functions, $r(s)$.

For the case of labeled trajectories, the values of $r(s)$ recovered by \ac{DM-IRL} are identical to the ground truth $r_{real}(s)$ since a regression model is used and we have the exact score for each user trajectory.

For the case of trajectories without scores, the quality of the recovered reward function is worse than \ac{DM-IRL}. \ac{MaxEnt-IRL} can only give a general overview of $r_{real}(s)$ if the user trajectories are optimal. If there are not enough constraints on the connections between states, with each iteration of running \ac{ISO}, the shape of the sampled trajectories becomes more similar, which means that most trajectories pass by the same states and the diversity of trajectories decreases. 
We found that this makes it even more difficult to recover $r_{real}(s)$ and the \ac{MaxEnt-IRL} quality deteriorates with the number of iterations, which results in lower performance in  Figure~\ref{fig:state_value_0}.

Hence, improving the performance of \ac{IRL} methods is likely to significantly boost the performance of \ac{ISO} and more advanced \ac{IRL} methods could be adopted according to the real task.

\subsection{Optimizing Interactive Systems in a Network-based World}
\label{sec:exp-neural}
\subsubsection{Experimental Setup}
\label{section:experimentalsetup_2}

\paragraph{\textbf{Designing an Interactive System with Neural Networks}}

\begin{figure*}[ht]
\centering
   \includegraphics[clip, width=0.8\textwidth]{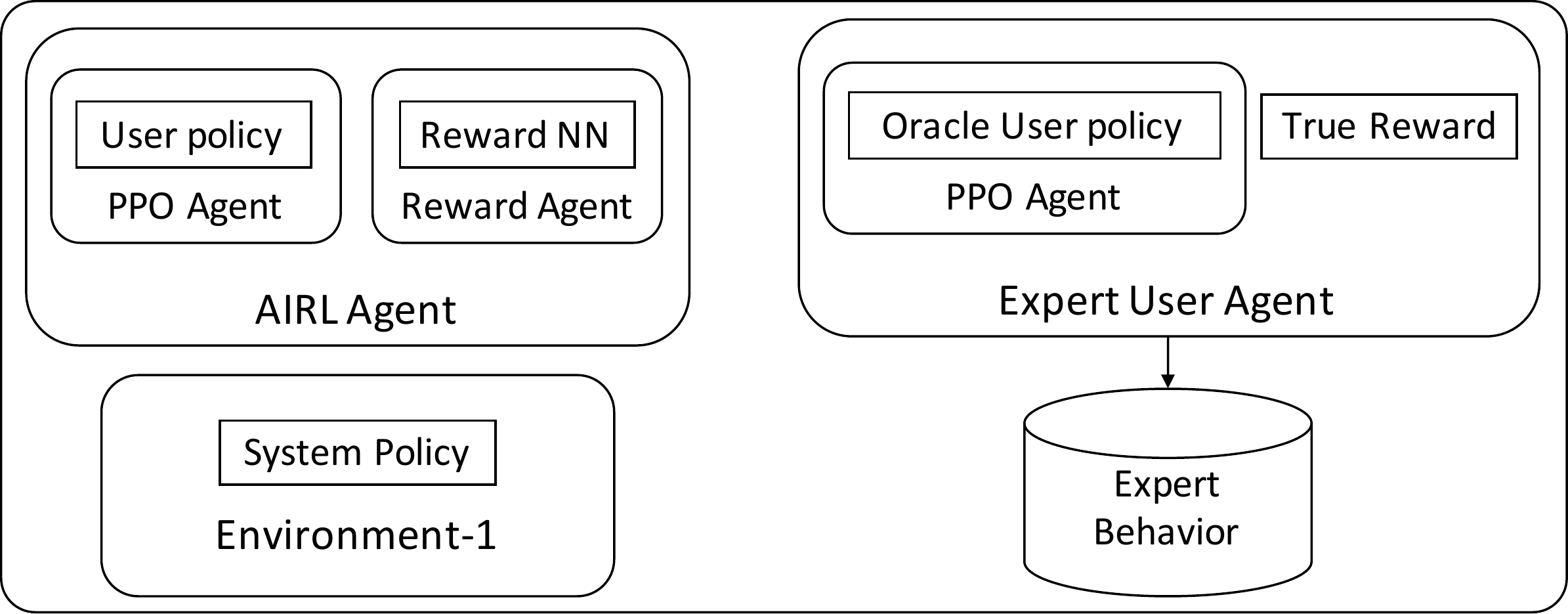}
  \caption{The architecture of the optimizing module in the original $\text{MDP}(S, A, T, r, \gamma)$, which is responsible of generating user behavior and recovering user reward functions.
  }
   \label{fig:mdp1}
\end{figure*}

\begin{figure*}[ht]
\centering
   \includegraphics[clip, width=0.5\textwidth]{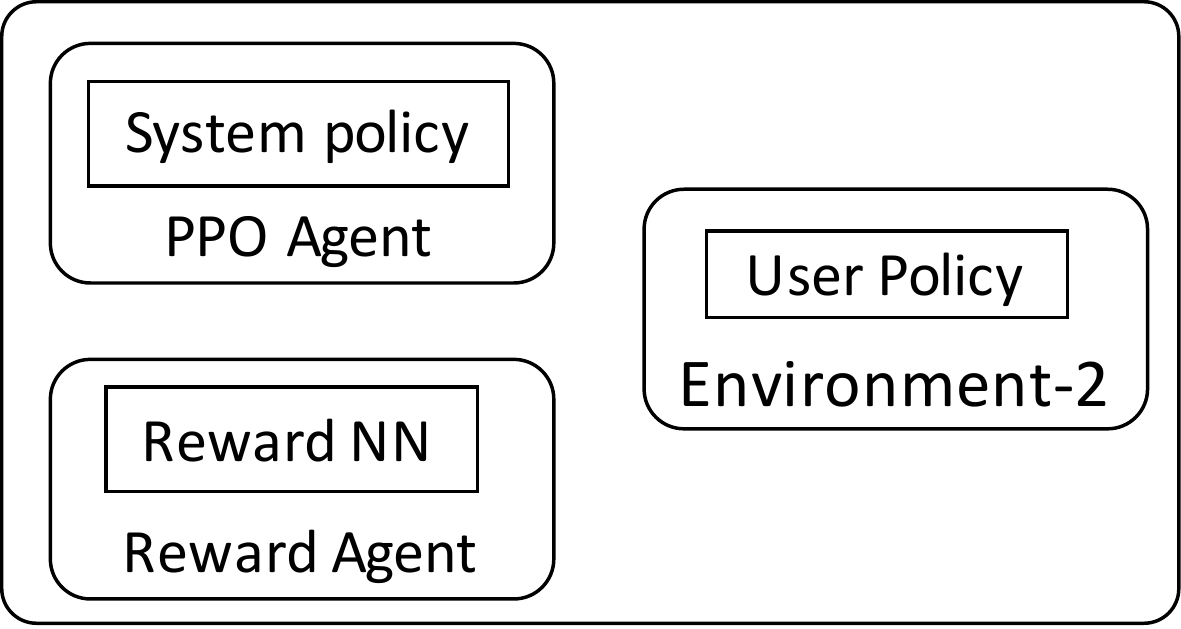}
   \caption{The architecture of the optimizing module in the reformulated $\text{MDP}^+(S^+, A^+, T^+, r^+, \gamma^+)$, responsible for optimizing the system agent.}
   \label{fig:mdp2}   
\end{figure*}

In this setup, we first present a simulated framework used for optimizing the \is $(S, A, T)$ with \ac{ISO}. Based on the two-step optimization setup in Section \ref{sec:opt-method} we designed two separate optimizing modules respectively. Figure~\ref{fig:mdp1} shows the architecture of the optimizing module for the original $\text{MDP}(S, A, T, r, \gamma)$, while Figure~\ref{fig:mdp2} describes the optimizing module for the reformulated $\text{MDP}^+(S^+, A^+, T^+, r^+, \gamma^+)$ respectively. As described in Section~\ref{sec:opt-method}, we use $\text{MDP}(S, A, T, r, \gamma)$ for reward learning and the reformulated $\text{MDP}^+(S^+, A^+, T^+, r^+, \gamma^+)$ for system optimization. \\

In the proposed setup, we have continuous state space $S$ and discrete action space $A$ for the original $\text{MDP}(S, A, T, r, \gamma)$, where the dimension of $S$ is $S_{dim}=50$ and action number is $|A|=10$. The user policy $\pi_{user}$,  the system policy $\pi_{sys}$ and the reward function $r(s)$ are represented with multi-layer perceptrons separately. Following Algorithm~\ref{alg:opt-env}, the transition distribution $T (S_{t+1} | S_t, A_t)$ is exactly the system policy $\pi_{sys}$ which is fixed in this step.  We assume the state distribution follows multivariate Gaussian distribution with a diagonal covariance matrix and the system policy $\pi_{sys}^+$ will produce the corresponding mean and variance. Since state space $S$ is continuous, the output of $\pi_{sys}$ will be a sampled continuous state $s_{t+1}$ at next step $t+1$ given $s_t$ and $a_t$. 
Here we use \ac{PPO}~\citep{schulman2017proximal}, a policy gradient based method, to optimize user policy $\pi_{user}$. With respect to the reformulated $\text{MDP}^+(S^+, A^+, T^+, r^+, \gamma^+)$, the state $s_t$ and action $a_t$ from the original $\text{MDP}$ will be concatenated to form the new state $s^+_t$ following Algorithm \ref{alg:opt-env}. The action $a^+$ is continuous and the transition distribution $T^+ (S_{t+1}^+ | S_t^+, A_t^+)$ is exactly the user policy $\pi_{user}$ in the original \text{MDP}. Different from $\pi_{sys}$ in the original \text{MDP}, $\pi_{sys}^+$ will be updated with PPO and it will be used to replace $\pi_{sys}$ in the original \text{MDP} after optimization finished. $r^+(s)$ is the learned reward function in the first optimizing step.

With respect to the optimizing module for $\text{MDP}(S, A, T, r, \gamma)$ shown in Figure~\ref{fig:mdp1}, the user policy is wrapped up with a \ac{PPO} agent and the reward function is loaded to the reward agent. To estimate the user reward function, we utilize \acf{AIRL} in the reward learning step. The user policy agent and the reward agent make up of the \ac{AIRL} agent. As an adversarial learning method, the \ac{AIRL} agent needs user traces generated by real user to update the reward function. In this setup, we use the user agent $\pi_{user\_real}^*$ trained with the true reward function $r_{real}(s)$ to produce necessary user-system interaction traces, which will be stored in the \textit{Expert Behavior} area. The environment \textit{Environment-1} for AIRL training and behavior generation mainly consists of the system policy $\pi_{sys}$ to deliver the next state $s_{t+1}$ given state $s_t$ and action $a_t$ according to $T(S_{t+1}|S_t, A_t)$. The system policy $\pi_{sys}$ will keep fixed in $\text{MDP}(S, A, T, r, \gamma)$. It should be noted that there are two different user reward functions in Figure \ref{fig:mdp1}. The reward function $r_{airl}(s)$ in AIRL agent is updated during AIRL training while the reward function $r_{real}(s)$ in the expert agent is the true reward function. The AIRL agent and system policy has no access to the true reward function $r_{real}(s)$ and we use $r_{airl}(s)$ to approximate $r_{real}(s)$, which is also the motivation of \ac{AIRL}. 

The optimizing module for the reformulated $\text{MDP}^+(S^+, A^+, T^+, r^+, \gamma^+)$ shown in Figure \ref{fig:mdp2} is responsible of updating the system policy $\pi_{sys}$with the recovered reward function $r_{airl}(s)$. Just like other reinforcement learning setups,  it mainly has three components: environment, PPO agent, and the reward function. The system policy $\pi_{sys}$ is wrapped up with a PPO agent and the reward agent is the function $r_{airl}(s)$ learned in $\text{MDP}(S, A, T, r, \gamma)$.   Given state $s^+_t$ and action $a^+_t$, the step function of the environment \textit{Environment-2} will return the next state $s^+_{t+1}$ according to $T^+ (S_{t+1}^+ | S_t^+, A_t^+)$ in Line 4 of Algorithm~\ref{alg:opt-env}, where the user policy $\pi_{user}$ is involved.

\paragraph{\textbf{Modeling User Behavior}}
Given the current system policy $\pi_{sys}$ and the real user reward function $r_{real}(s)$, we optimize the user policy $\pi_{user}$ by running \ac{PPO} method. The optimized user policy will be saved as the oracle user policy $\pi_{user\_real}^*$. Then by making the user policy $\pi_{user\_real}^*$ interact with the system policy $\pi_{sys}$, we can collect a bunch of interaction trajectories ($20K$ in our experiments) and all these behavior data will be loaded to the expert behavior bucket. The maximum length of collected trajectories is $40$.  The stored user interaction traces will be used to learn the user reward function $r_{airl}(s)$ (we use \ac{AIRL} method in this setup).

\paragraph{\textbf{Evaluation Process}}
To evaluate the performance of \ac{ISO} in the proposed framework, we report the \textit{Average Return} of $m$ sampled trajectories ($m=1000$ in our setup) under optimal policy $\pi_{user\_real}^*$ under the real reward $r_{real}(s)$ for an \emph{initial} \is and an \emph{optimized} one, which we derive after 3 \textit{iterations} (one \textit{iteration} means we sequentially recover the reward function and run Algorithm \ref{alg:opt-env} once). 

A higher average return means users are more satisfied while interacting with the \is. We initialize $5$ different initial 
\iss by randomly sampling the system policy $\pi_{sys}$, and report the overall performance over these $5$ systems. Besides, we want to avoid the situations that the optimized system has totally different behaviors compared to the initial system because the dramatic change may hurt users' experience. To make sure the systems before and after optimized follow similar behaviors, we introduce a regularization term, the KL-Divergence $\lambda * D_{KL}(T_{opt} \| T_{init})$,  to control the distance between these two system policies. This term can also be regarded as the inherent constraints between state transitions, just like the ``connection factor'' in Section \ref{section:experimentalsetup}. The hyperparameter $\lambda$ is applied to control the affect of the term. Due to the training complexity of network-based simulations,  we run $5$ times for each parameter setup rather than $40$ times in the tabular world.
\paragraph{\textbf{The Ground Truth of User Reward Functions}}
With respect to the true reward function $r_{real}$, we have two different setups: \textit{a handcrafted reward function} and \textit{a randomly initialized reward function}. For the handcrafted one, we use $r_{real}(s)=\frac{1}{s_{dim}}\phi(s) * \phi(s)$ as the reward for the given state $s$. In terms of the sampled reward function, we initialize the parameters of the reward network with uniform distributions, and this makes recovering the reward function more difficult because there are no patterns in the sampled reward function. In the real world, users always have their preferences and the reward function in users’ minds is not likely to be random. The true reward function $r_{real}(s)$ is fixed in the whole optimizing process.

\subsubsection{Results and Discussion}
\label{section:experimentalresult_2}
In this section,  we first discuss the results of the experiments with a manually designed real reward function. Then, we move to the discussion of the experimental results with randomly initialized reward function.

\begin{figure*}[ht!]
\centering
   \includegraphics[clip, width=0.8\textwidth]{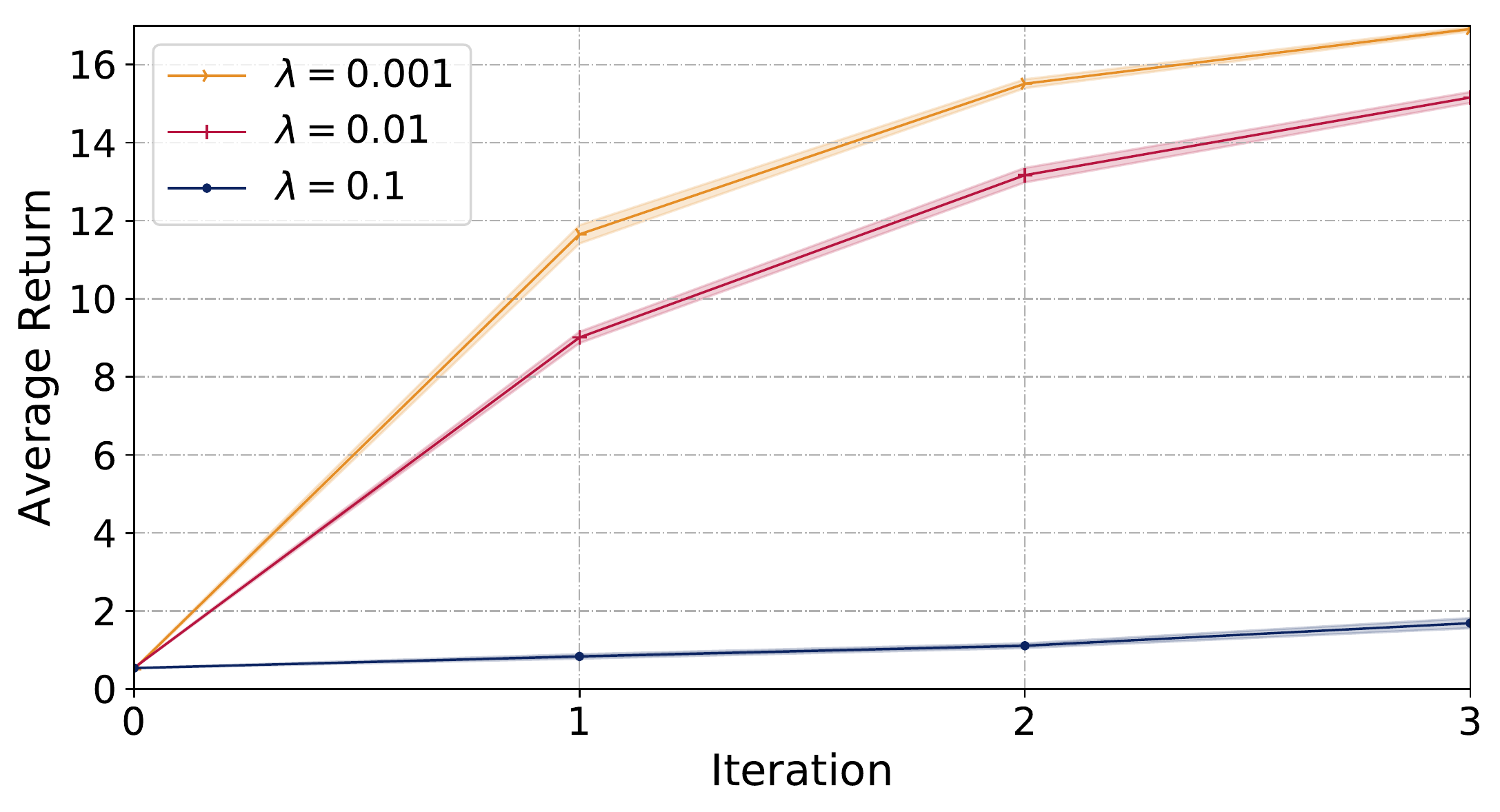}
   \caption{The state performance during optimization with oracle reward function and oracle user policy. The real reward function is manually designed. The error bounds denote the standard error of the mean ($\pm$SEM). 
   }
   \label{fig:oracle_pi_oracle_reward}{}
\end{figure*}

\begin{figure*}[ht!]
\centering
   \includegraphics[clip, width=0.8\textwidth]{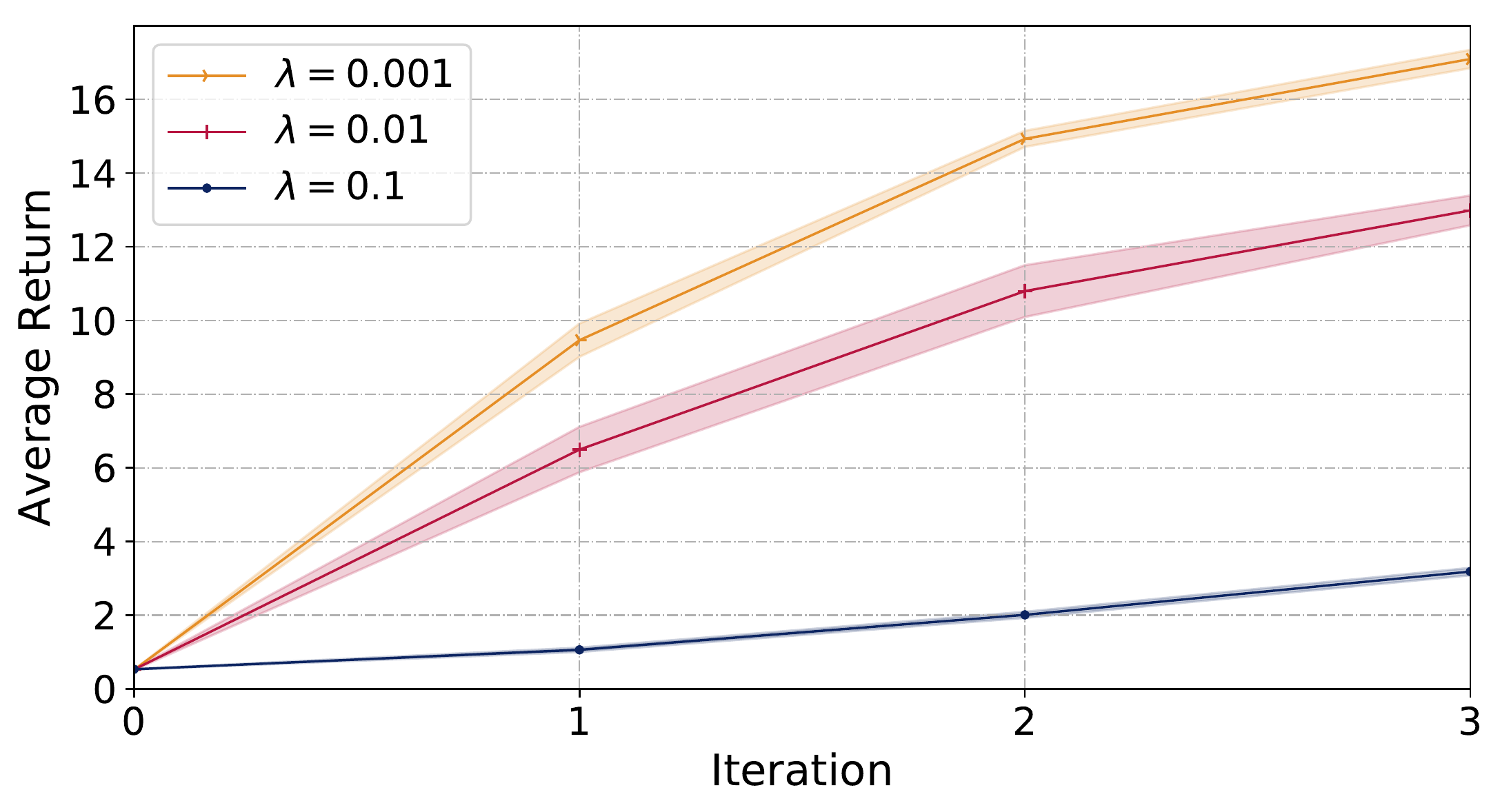}
   \caption{The state performance during optimization with recovered reward and oracle user policy. The real reward function is manually designed. The error bounds denote the standard error of the mean ($\pm$SEM).}
   \label{fig:oracle_pi_airl_reward}
\end{figure*}

\begin{figure*}[ht!]
\centering
   \includegraphics[clip, width=0.8\textwidth]{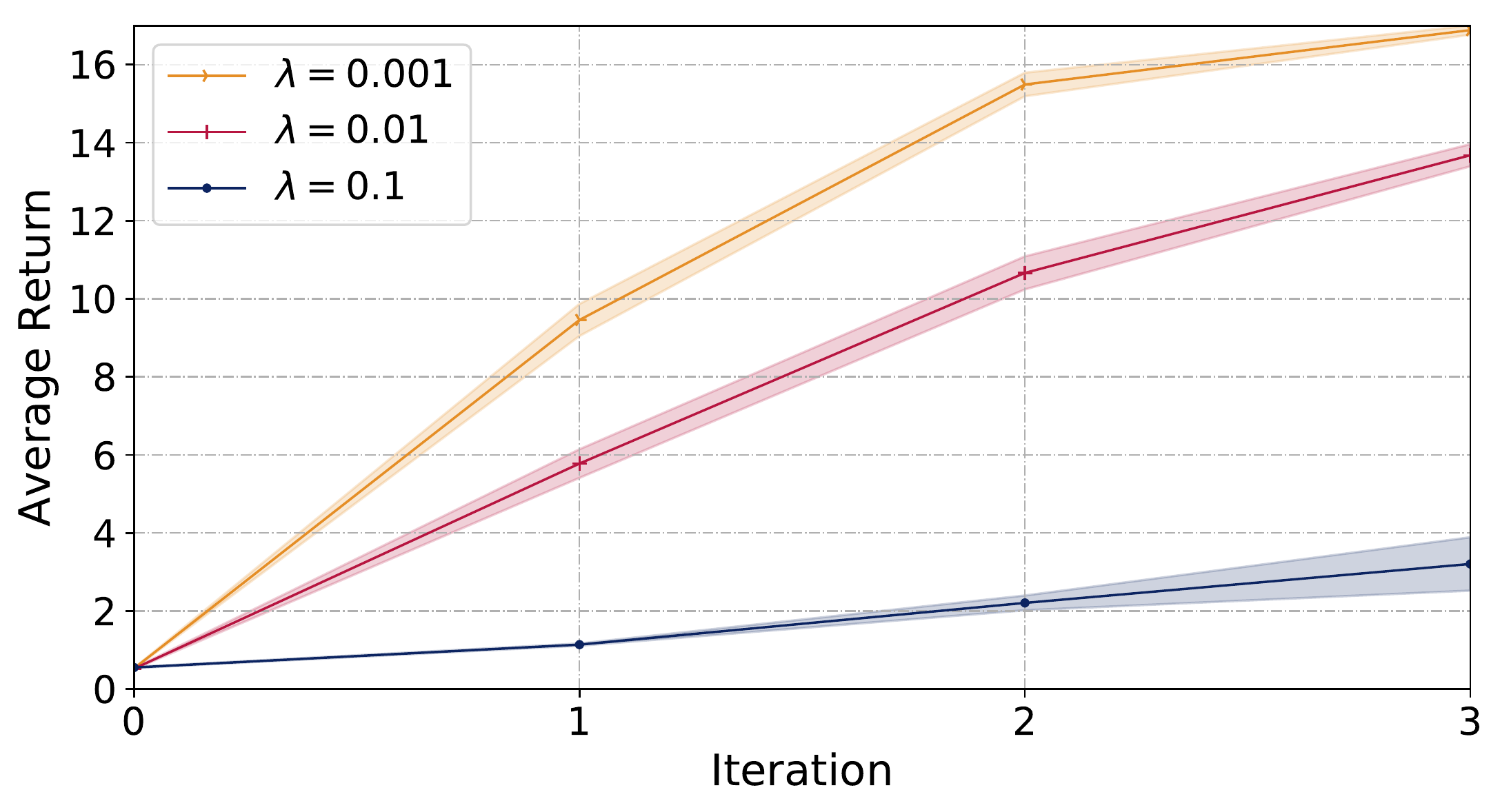}
   \caption{The state performance during optimization with recovered reward function and recovered user policy. The real reward function is manually designed. The error bounds denote the standard error of the mean ($\pm$SEM).}
   \label{fig:airl_pi_airl_reward}
\end{figure*}

\paragraph{\textbf{Manually Designed Real Reward Function}}
To verify if the proposed two-MDP framework works or not, we first skip the reward learning step and use the oracle reward function $r_{real}(s)$ as the ``learned'' reward function with collected user behaviors. With respect to the user policy $\pi_{user}$ used to interact with the system agent in the second optimizing module, we use the oracle user policy $\pi_{user\_real}^*$ trained with true reward function $r_{real}(s)$. Other modules keep the same and we obtain the performance in Figure \ref{fig:oracle_pi_oracle_reward}. An interactive sytem at iteration $0$ is the initial system and not optimized yet.
As we can see, with a looser restriction (i.e., a smaller $\lambda$ value) on the distance between the optimized system and the initial system, we can achieve higher performance with respect to the average trajectory returns. 
After we bring back the reward learning step and use the learned reward function $r_{airl}(s)$ to optimize the system policy, we have the results shown  in Figure \ref{fig:oracle_pi_airl_reward}. The system can still achieve higher performance by running Algorithm \ref{alg:opt-env}. If we compare the results between systems $\lambda=0.001$ in Figure \ref{fig:oracle_pi_oracle_reward} and  Figure \ref{fig:oracle_pi_airl_reward}, we can find that the system trained with oracle reward $r_{real}(s)$ can hit higher returns after two iterations. The finding still holds with respect to the systems $\lambda=0.01$ in both setups. However, this is not the case when we set $\lambda=0.1$. We suspect this is because the large regularization term $D_{KL}(T_{opt} \| T_{init})$ has brought too many uncontrollable factors into the optimization step, which may disturb the training. 

As mentioned in Section \ref{section:experimentalsetup_2}, the user policy $\pi_{user}$ is essential while optimizing the system $\pi_{sys}$. In Algorithm \ref{alg:opt-env}, the user policy $\pi_{user}$ plays the role of the transition distribution $T^+$ in the environment \textit{Environment-2}. In addition to the two reward function setups above, we need to conduct an experiment with the user policy $\pi_{user}$ trained with recovered reward function $r_{airl}(s)$ for system optimization in the reformulated $\text{MDP}^+$.
Since we use AIRL to learn the reward function in this framework, we have the estimated user policy $\pi_{user\_airl}^*$ which is rebuilt during the adversarial training process. We replace $\pi_{user\_real}^*$ in the environment \textit{Environment-2} with this rebuilt policy $\pi_{user\_airl}^*$. In terms of the reward function $r^+$ in $\text{MDP}$$^+(S^+, A^+, T^+, r^+, \gamma^+)$, we use the reward function $r_{airl}(s)$. By running Algorithm \ref{alg:opt-env}, we have the results in Figure \ref{fig:airl_pi_airl_reward}. It is clear that the rebuilt policy $\pi_{user\_airl}^*$ can still help with improving the system performance. This is meaningful because by using adversarial training we can rebuilt the user policy and user reward function simultaneously. The accuracy of the estimated user policy will definitely benefit from a high quality estimation of the user reward function. The only moment that real users are involved happens when we are collecting user-system interaction trajectories. This perfectly matches the scenarios in real life, where we first collect interaction histories from users and then infer the user preferences ($r_{airl}$) and user behavior patterns ($\pi_{user\_airl}$) according to the collected data. In the next step, the system policy $\pi_{sys}$ will be optimized based on user preferences and user behavior patterns. In the end, the newly updated system $(S, A, T^*)$ will be presented to users to improve their user experience. If necessary, new interaction trajectories will be collected and another optimization turn can start again.

\begin{figure*}[ht!]
\centering
   \includegraphics[clip, width=0.8\textwidth]{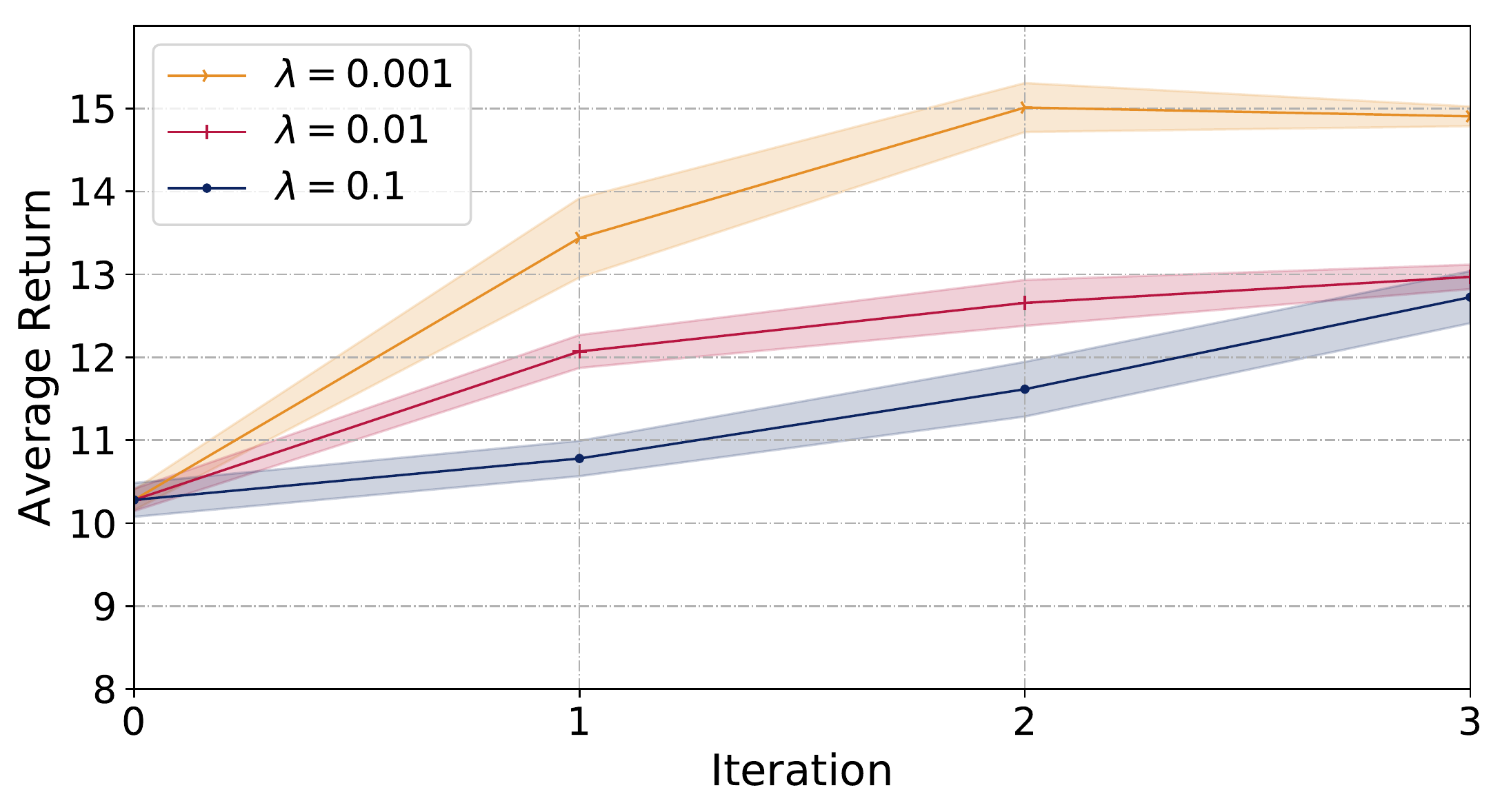}
   \caption{The state performance during optimization with oracle reward function and oracle user policy. The real reward function is randomly initialized. The error bounds denote the standard error of the mean ($\pm$SEM).}
   \label{fig:oracle_pi_oracle_reward_rd}
\end{figure*}

\begin{figure*}[ht!]
\centering
   \includegraphics[clip, width=0.8\textwidth]{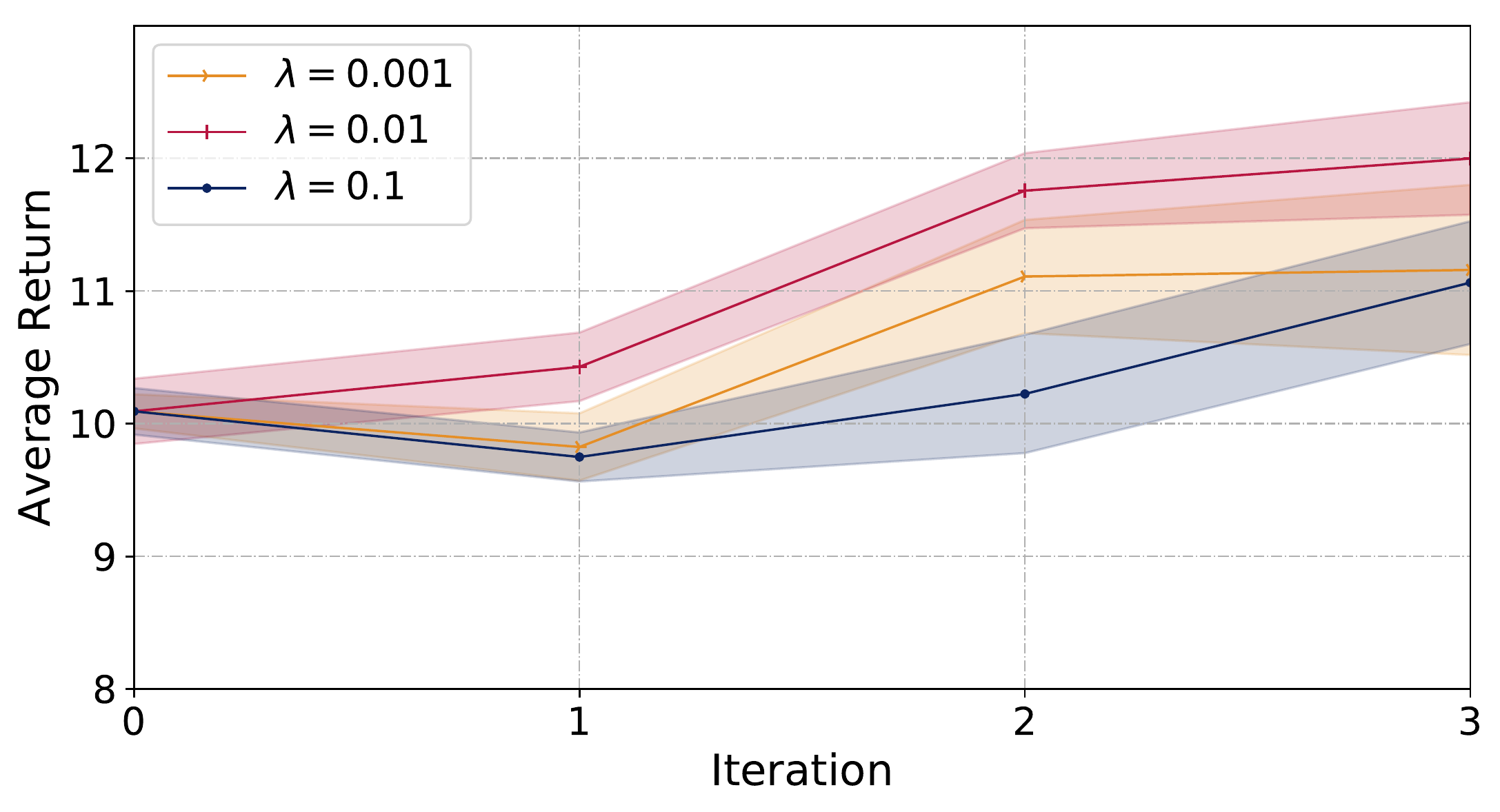}
   \caption{The state performance during optimization with recovered reward and oracle user policy. The real reward function is randomly initialized. The error bounds denote the standard error of the mean ($\pm$SEM).}
   \label{fig:oracle_pi_airl_reward_rd}
\end{figure*}

\begin{figure*}[ht!]
\centering
   \includegraphics[clip, width=0.8\textwidth]{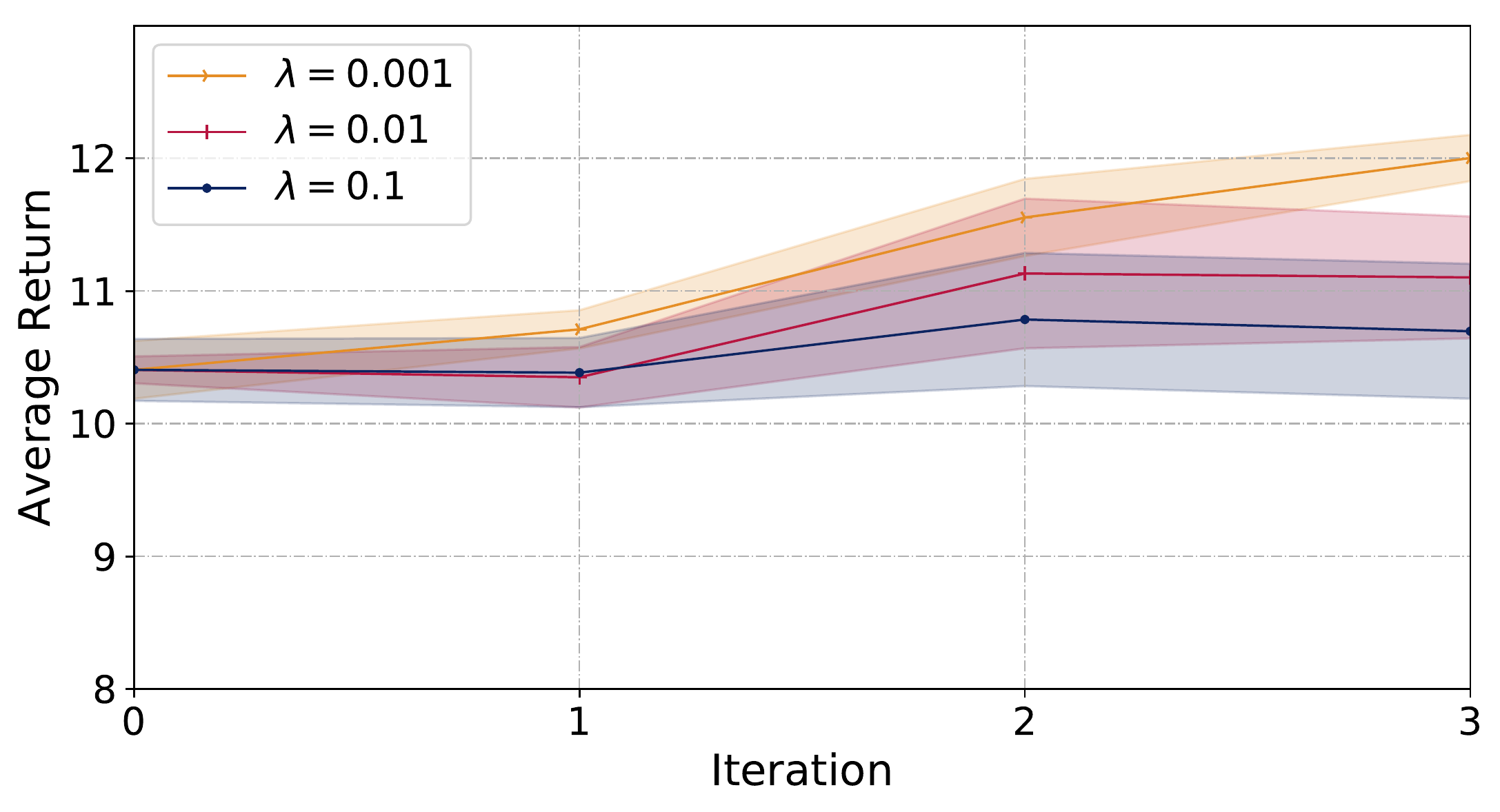}
   \caption{The state performance during optimization with recovered reward function and recovered user policy. The real reward function is randomly initialized. The error bounds denote the standard error of the mean ($\pm$SEM).}
   \label{fig:airl_pi_airl_reward_rd}
\end{figure*}

\paragraph{\textbf{Randomly Initialized Reward Function}} In this section, we show how the interactive optimizer performs when the reward function $r_{real}(s)$ is randomly initialized. In Figure~\ref{fig:oracle_pi_oracle_reward_rd}, with the real reward function $r_{real}(s)$, the system can still achieve relatively large improvements in terms of average return. All curves have higher starting points is because the randomly initialized system policy has advantage to hit higher reward for a random reward function and this will not hold when the reward function has special pattern like Section \ref{section:experimentalresult_2}. We also find that looser restrictions on the distance between the optimized system and the initial system can bring larger performance improvements, as we observed in  Section \ref{section:experimentalresult_2}.

 With respect to Figure~\ref{fig:oracle_pi_airl_reward_rd} and Figure \ref{fig:airl_pi_airl_reward_rd}, the improvements still exist but are not so significant compared to those with the handcrafted reward function in Figure \ref{fig:oracle_pi_airl_reward} and Figure \ref{fig:airl_pi_airl_reward}. The potential reason is that it is hard to recover a high quality reward function given user behaviors generated by a random reward function. Especially before the first iteration, the system still performs randomly (the initial system is randomly initialized) and it is difficult to collect useful interaction traces for reward learning, and this is also the reason why, in Figure \ref{fig:oracle_pi_airl_reward_rd} and Figure \ref{fig:airl_pi_airl_reward_rd}, the average returns of some curves even drop after the first iteration. However, in the real world, users always have their preferences and the reward function in users' minds is unlikely to be random. Besides, the initial system will not behave randomly because in most cases a real interactive system (e.g., search engine, digital assistant) will be tested offline first and will not be deployed before it can achieve reasonable performance. These will alleviate the reward learning stress at some degree. We have this random reward function here simply to validate how well the method could perform with most uncontrollable behaviors.

\subsection{Limitations (as Future Directions)}
\label{sec:limitations}

First, to recover a reliable reward function, a large number of high-quality user interaction traces are essential, which can come with a great cost in real life (but not impossible). Furthermore, an interactive system usually serves different users, which can lead to the violation of~\ref{ass:user-behave} about the homogeneity of user behavior. Therefore, we would need to work on the personalizing of the recovered reward functions. A possible way to address this limitation in the future is to incorporate the user features into the state space, but this still needs to be explored.

Second, as shown in Section~\ref{sec:results}, the final performance of the optimized system highly relies on the quality of the recovered reward function. With respect to the more advanced extension of \acf{MaxEnt-IRL}, which is \acf{AIRL}, the adversarial training process is intractable for complex real behaviors. 
The two limitations above will boil down to the quality of recovered reward functions, given limited user traces in real life. 

Third, after we have inferred the reward function, we will update the system in a reformulated MDP setup, where we switch the roles between the agent and environment. The potential problem that can arise is that the action space for the new MDP could be extremely large and this may present challenges for the scalability of the \acf{RL} process.

Finally, we validate our method in two simulated experimental setups. Despite the fact that we try to design our setups as close as possible to the real-word scenarios, there is a potential gap between the designed systems and real-world applications. But the positive verification of our method in simulated setup helps us to understand better the pros and cons of the proposed approach and help us with planning the experiments with the real-world scenarios in the near future.

\medskip
To summarize, we have proposed two experimental setups to test the proposed framework: tabular and neural. In both cases, the results demonstrate significant improvements of \iss when applying our method. We conclude this section acknowledging a number of possible limitations, some of which can be considered as future directions.

\section{Conclusions and Future Work}
\label{section:conclusion}

We have recognized that previous work on \iss has relied on the assumption that the handcrafted objective functions can accurately reflect users' preferences and intentions while interacting with interaction systems. 
As a result, \iss have been optimized for manually designed objectives that do not always align with the true user preferences and cannot be generalized across different domains.
To overcome this discrepancy, we have proposed a novel two-step framework to optimize interactive systems, which first infer the user reward model given collected user interaction traces and then update the system with the inferred reward functions via a novel algorithm: the \acf{ISO}.

Firstly, we modeled user-system interactions using \ac{MDP}, where the agent is the user, and the stochastic environment is the \is. 
User satisfaction is modeled via rewards received from interactions, and the user interaction history is represented by a set of trajectories.
We followed the previously justified assumption that user incentive to interact with the system if they are rewarded. Treating an interactive system as a changeable and programmable environment is novel and reasonable because we have complete control of the interactive systems since we are the system designers. 

Secondly, we formalized an optimization problem to infer the user needs from the observed user-system interactions, in the form of a data-driven objective. Importantly, our method works without any domain knowledge, and is thus even applicable when prior knowledge is absent.

Thirdly, we proposed a novel, \acf{ISO}, that iterates between optimizing the \is for the current inferred objective; and letting the user adapt to the new system behavior.
This process repeats until both the user and system policies converge. Our experimental results show that \ac{ISO} robustly improves the user satisfaction.

The newly proposed approach to optimize an \is based on data-driven objectives is novel, many promising directions for future work are possible.
For instance, while \ac{ISO} performs well for users with a single goal, this approach could be extended to settings with multiple goals.
Similarly, extensions considering more personalized goals could benefit the overall user experience.
Finally, investigating the scalability and real world applicability of \ac{ISO} could open many research possibilities.


\bibliography{bibliography}   

\begin{thebibliography}{101}
\providecommand{\natexlab}[1]{#1}
\providecommand{\url}[1]{\texttt{#1}}
\expandafter\ifx\csname urlstyle\endcsname\relax
  \providecommand{\doi}[1]{doi: #1}\else
  \providecommand{\doi}{doi: \begingroup \urlstyle{rm}\Url}\fi

\bibitem[Abbeel and Ng(2004)]{abbeel_icml_2004}
P.~Abbeel and A.~Y. Ng.
\newblock Apprenticeship learning via inverse reinforcement learning.
\newblock In \emph{ICML}, pages 1--8. ACM, 2004.

\bibitem[Akkaya et~al.(2019)Akkaya, Andrychowicz, Chociej, Litwin, McGrew,
  Petron, Paino, Plappert, Powell, Ribas, et~al.]{akkaya2019solving}
I.~Akkaya, M.~Andrychowicz, M.~Chociej, M.~Litwin, B.~McGrew, A.~Petron,
  A.~Paino, M.~Plappert, G.~Powell, R.~Ribas, et~al.
\newblock Solving rubik's cube with a robot hand.
\newblock \emph{arXiv preprint arXiv:1910.07113}, 2019.

\bibitem[Argelaguet et~al.(2016)Argelaguet, Hoyet, Trico, and
  L{\'e}cuyer]{argelaguet2016role}
F.~Argelaguet, L.~Hoyet, M.~Trico, and A.~L{\'e}cuyer.
\newblock The role of interaction in virtual embodiment: Effects of the virtual
  hand representation.
\newblock In \emph{VR}, pages 3--10. IEEE, 2016.

\bibitem[Azzopardi(2014)]{azzopardi_sigir_2014}
L.~Azzopardi.
\newblock Modelling interaction with economic models of search.
\newblock In \emph{SIGIR}, pages 3--12. ACM, 2014.

\bibitem[Blume et~al.(2008)Blume, Durlauf, and Blume]{blume2008new}
L.~E. Blume, S.~Durlauf, and L.~E. Blume.
\newblock \emph{The new Palgrave dictionary of economics}.
\newblock Palgrave Macmillan Manchester, 2008.

\bibitem[Borisov et~al.(2016)Borisov, Markov, de~Rijke, and
  Serdyukov]{borisov_www_2016}
A.~Borisov, I.~Markov, M.~de~Rijke, and P.~Serdyukov.
\newblock A neural click model for web search.
\newblock In \emph{WWW}, pages 531--541, 2016.

\bibitem[Boularias et~al.(2011)Boularias, Kober, and
  Peters]{boularias2011relative}
A.~Boularias, J.~Kober, and J.~Peters.
\newblock Relative entropy inverse reinforcement learning.
\newblock In \emph{AISTATS}, pages 182--189, 2011.

\bibitem[Burchfiel et~al.(2016)Burchfiel, Tomasi, and
  Parr]{Burchfiel_aaai_2016}
B.~Burchfiel, C.~Tomasi, and R.~Parr.
\newblock Distance minimization for reward learning from scored trajectories.
\newblock In \emph{{AAAI}}, pages 3330--3336. {AAAI} Press, 2016.

\bibitem[Chen et~al.(2019)Chen, Beutel, Covington, Jain, Belletti, and
  Chi]{chen2019top}
M.~Chen, A.~Beutel, P.~Covington, S.~Jain, F.~Belletti, and E.~H. Chi.
\newblock Top-k off-policy correction for a reinforce recommender system.
\newblock In \emph{WSDM}, pages 456--464. ACM, 2019.

\bibitem[Christiano et~al.(2017)Christiano, Leike, Brown, Martic, Legg, and
  Amodei]{christiano2017deep}
P.~F. Christiano, J.~Leike, T.~Brown, M.~Martic, S.~Legg, and D.~Amodei.
\newblock Deep reinforcement learning from human preferences.
\newblock In \emph{Advances in Neural Information Processing Systems}, pages
  4299--4307, 2017.

\bibitem[Cui et~al.(2019)Cui, Wang, Song, Huang, Xu, and Nie]{cui2019user}
C.~Cui, W.~Wang, X.~Song, M.~Huang, X.-S. Xu, and L.~Nie.
\newblock User attention-guided multimodal dialog systems.
\newblock In \emph{SIGIR}, pages 445--454. ACM, 2019.

\bibitem[Dehghani et~al.(2017)Dehghani, Zamani, Severyn, Kamps, and
  Croft]{dehghani2017neural}
M.~Dehghani, H.~Zamani, A.~Severyn, J.~Kamps, and W.~B. Croft.
\newblock Neural ranking models with weak supervision.
\newblock In \emph{SIGIR}, pages 65--74, 2017.

\bibitem[Dhingra et~al.(2016)Dhingra, Li, Li, Gao, Chen, Ahmed, and
  Deng]{dhingra2016towards}
B.~Dhingra, L.~Li, X.~Li, J.~Gao, Y.-N. Chen, F.~Ahmed, and L.~Deng.
\newblock Towards end-to-end reinforcement learning of dialogue agents for
  information access.
\newblock \emph{arXiv preprint arXiv:1609.00777}, 2016.

\bibitem[Drutsa et~al.(2015)Drutsa, Gusev, and Serdyukov]{Drutsa_wsdm_2015}
A.~Drutsa, G.~Gusev, and P.~Serdyukov.
\newblock Engagement periodicity in search engine usage: Analysis and its
  application to search quality evaluation.
\newblock In \emph{WSDM}, pages 27--36, 2015.

\bibitem[Duan et~al.(2016)Duan, Schulman, Chen, Bartlett, Sutskever, and
  Abbeel]{duan2016rl}
Y.~Duan, J.~Schulman, X.~Chen, P.~L. Bartlett, I.~Sutskever, and P.~Abbeel.
\newblock ${RL}^2$: Fast reinforcement learning via slow reinforcement
  learning.
\newblock \emph{arXiv preprint arXiv:1611.02779}, 2016.

\bibitem[Dupret and Lalmas(2013)]{Dupret_wsdm_2013}
G.~Dupret and M.~Lalmas.
\newblock Absence time and user engagement: evaluating ranking functions.
\newblock In \emph{WSDM}, pages 173--182, 2013.

\bibitem[El~Asri et~al.(2013)El~Asri, Laroche, and Pietquin]{el2013_dm_irl}
L.~El~Asri, R.~Laroche, and O.~Pietquin.
\newblock Reward shaping for statistical optimisation of dialogue management.
\newblock In \emph{SLSP}, pages 93--101. Springer, 2013.

\bibitem[Finn et~al.(2016{\natexlab{a}})Finn, Christiano, Abbeel, and
  Levine]{finn2016connection}
C.~Finn, P.~Christiano, P.~Abbeel, and S.~Levine.
\newblock A connection between generative adversarial networks, inverse
  reinforcement learning, and energy-based models.
\newblock \emph{arXiv preprint arXiv:1611.03852}, 2016{\natexlab{a}}.

\bibitem[Finn et~al.(2016{\natexlab{b}})Finn, Levine, and
  Abbeel]{guided_cost_learning}
C.~Finn, S.~Levine, and P.~Abbeel.
\newblock Guided cost learning: Deep inverse optimal control via policy
  optimization.
\newblock In \emph{ICML}, pages 49--58, 2016{\natexlab{b}}.

\bibitem[Finn et~al.(2017)Finn, Yu, Zhang, Abbeel, and Levine]{finn_corl17}
C.~Finn, T.~Yu, T.~Zhang, P.~Abbeel, and S.~Levine.
\newblock One-shot visual imitation learning via meta-learning.
\newblock In \emph{CoRL}, pages 357--368, 2017.

\bibitem[Fox et~al.(2005)Fox, Karnawat, Mydland, Dumais, and
  White]{Fox_trans_2005}
S.~Fox, K.~Karnawat, M.~Mydland, S.~T. Dumais, and T.~White.
\newblock Evaluating implicit measures to improve web search.
\newblock \emph{ACM Transactions on Information Systems}, 23\penalty0
  (2):\penalty0 147--168, 2005.

\bibitem[Fu et~al.(2017)Fu, Luo, and Levine]{fu2017learning}
J.~Fu, K.~Luo, and S.~Levine.
\newblock Learning robust rewards with adversarial inverse reinforcement
  learning.
\newblock \emph{arXiv preprint arXiv:1710.11248}, 2017.

\bibitem[Haarnoja et~al.(2018)Haarnoja, Zhou, Abbeel, and
  Levine]{haarnoja2018soft}
T.~Haarnoja, A.~Zhou, P.~Abbeel, and S.~Levine.
\newblock Soft actor-critic: Off-policy maximum entropy deep reinforcement
  learning with a stochastic actor.
\newblock \emph{arXiv preprint arXiv:1801.01290}, 2018.

\bibitem[Hafner et~al.(2019)Hafner, Lillicrap, Ba, and
  Norouzi]{hafner2019dream}
D.~Hafner, T.~Lillicrap, J.~Ba, and M.~Norouzi.
\newblock Dream to control: Learning behaviors by latent imagination.
\newblock \emph{arXiv preprint arXiv:1912.01603}, 2019.

\bibitem[Ho and Ermon(2016)]{ho2016generative}
J.~Ho and S.~Ermon.
\newblock Generative adversarial imitation learning.
\newblock In \emph{Advances in neural information processing systems}, pages
  4565--4573, 2016.

\bibitem[Hofmann et~al.(2011)Hofmann, Whiteson, and de~Rijke]{hofmann2011_ecir}
K.~Hofmann, S.~Whiteson, and M.~de~Rijke.
\newblock Balancing exploration and exploitation in learning to rank online.
\newblock In \emph{ECIR}, pages 251--263. Springer, 2011.

\bibitem[Hofmann et~al.(2013{\natexlab{a}})Hofmann, Schuth, Whiteson, and
  de~Rijke]{hofmann2013_wsdm}
K.~Hofmann, A.~Schuth, S.~Whiteson, and M.~de~Rijke.
\newblock Reusing historical interaction data for faster online learning to
  rank for {IR}.
\newblock In \emph{WSDM}, pages 183--192. ACM, 2013{\natexlab{a}}.

\bibitem[Hofmann et~al.(2013{\natexlab{b}})Hofmann, Whiteson, and
  de~Rijke]{hofmann-balancing-2013}
K.~Hofmann, S.~Whiteson, and M.~de~Rijke.
\newblock Balancing exploration and exploitation in listwise and pairwise
  online learning to rank for information retrieval.
\newblock \emph{Information Retrieval Journal}, 16\penalty0 (1):\penalty0
  63--90, 2013{\natexlab{b}}.

\bibitem[J{\"a}rvelin and Kek{\"a}l{\"a}inen(2002)]{Kalervo_2002}
K.~J{\"a}rvelin and J.~Kek{\"a}l{\"a}inen.
\newblock Cumulated gain-based evaluation of ir techniques.
\newblock \emph{ACM Transactions on Information Systems}, 20\penalty0
  (4):\penalty0 422--446, 2002.

\bibitem[Jeon et~al.(2020)Jeon, Milli, and Dragan]{jeon2020reward}
H.~J. Jeon, S.~Milli, and A.~D. Dragan.
\newblock Reward-rational (implicit) choice: A unifying formalism for reward
  learning.
\newblock \emph{arXiv preprint arXiv:2002.04833}, 2020.

\bibitem[Joachims et~al.(2005)Joachims, Granka, Pan, Hembrooke, and
  Gay]{Joachims_2005}
T.~Joachims, L.~Granka, B.~Pan, H.~Hembrooke, and G.~Gay.
\newblock Accurately interpreting clickthrough data as implicit feedback.
\newblock In \emph{SIGIR}, pages 154--161, 2005.

\bibitem[Kelly(2009)]{kelly-methods-2009}
D.~Kelly.
\newblock Methods for evaluating interactive information retrieval systems with
  users.
\newblock \emph{Foundations and Trends in Information Retrieval}, 3\penalty0
  (1--2):\penalty0 1--224, 2009.

\bibitem[Kelly(2015)]{kelly2015effort}
D.~Kelly.
\newblock When effort exceeds expectations: A theory of search task difficulty.
\newblock In \emph{ECIR Supporting Complex Search Task Workshop ‘15}, 2015.

\bibitem[Kiseleva and de~Rijke(2017)]{kiseleva2017evaluating}
J.~Kiseleva and M.~de~Rijke.
\newblock Evaluating personal assistants on mobile devices.
\newblock \emph{arXiv preprint arXiv:1706.04524}, 2017.

\bibitem[Kiseleva et~al.(2014)Kiseleva, Crestan, Brigo, and
  Dittel]{kiseleva_cikm_2014}
J.~Kiseleva, E.~Crestan, R.~Brigo, and R.~Dittel.
\newblock Modelling and detecting changes in user satisfaction.
\newblock In \emph{CIKM}, pages 1449--1458, 2014.

\bibitem[Kiseleva et~al.(2015)Kiseleva, Kamps, Nikulin, and
  Makarov]{kiseleva_serp_sigir_2015}
J.~Kiseleva, J.~Kamps, V.~Nikulin, and N.~Makarov.
\newblock Behavioral dynamics from the serp's perspective: What are failed
  serps and how to fix them?
\newblock In \emph{Submission of SIGIR}, 2015.

\bibitem[Kiseleva et~al.(2016{\natexlab{a}})Kiseleva, Williams, Awadallah,
  Zitouni, Crook, and Anastasakos]{kiseleva_sigir_2016}
J.~Kiseleva, K.~Williams, A.~H. Awadallah, I.~Zitouni, A.~Crook, and
  T.~Anastasakos.
\newblock Predicting user satisfaction with intelligent assistants.
\newblock In \emph{SIGIR}, pages 45--54. ACM, 2016{\natexlab{a}}.

\bibitem[Kiseleva et~al.(2016{\natexlab{b}})Kiseleva, Williams, Jiang,
  Awadallah, Zitouni, Crook, and Anastasakos]{kiseleva_chiir_2016}
J.~Kiseleva, K.~Williams, J.~Jiang, A.~H. Awadallah, I.~Zitouni, A.~Crook, and
  T.~Anastasakos.
\newblock Understanding user satisfaction with intelligent assistants.
\newblock In \emph{CHIIR}, pages 121--130, 2016{\natexlab{b}}.

\bibitem[Kosinski et~al.(2013)Kosinski, Stillwell, and Graepe]{Kosinski_2013}
M.~Kosinski, D.~Stillwell, and T.~Graepe.
\newblock Private traits and attributes are predictable from digital records of
  human behavior.
\newblock \emph{PNAS}, 110:\penalty0 5802--5805, 2013.

\bibitem[Kutlu et~al.(2018)Kutlu, Khetan, and Lease]{kutlu2018correlation}
M.~Kutlu, V.~Khetan, and M.~Lease.
\newblock Correlation and prediction of evaluation metrics in information
  retrieval.
\newblock \emph{arXiv preprint arXiv:1802.00323}, 2018.

\bibitem[Leike et~al.(2018)Leike, Krueger, Everitt, Martic, Maini, and
  Legg]{leike2018scalable}
J.~Leike, D.~Krueger, T.~Everitt, M.~Martic, V.~Maini, and S.~Legg.
\newblock Scalable agent alignment via reward modeling: a research direction.
\newblock \emph{arXiv preprint arXiv:1811.07871}, 2018.

\bibitem[Levine et~al.(2016{\natexlab{a}})Levine, Finn, Darrell, and
  Abbeel]{levine2016end2end}
S.~Levine, C.~Finn, T.~Darrell, and P.~Abbeel.
\newblock End-to-end training of deep visuomotor policies.
\newblock \emph{Journal of Machine Learning Research}, 17\penalty0
  (39):\penalty0 1--40, 2016{\natexlab{a}}.

\bibitem[Levine et~al.(2016{\natexlab{b}})Levine, Pastor, Krizhevsky, Ibarz,
  and Quillen]{levine2016learning}
S.~Levine, P.~Pastor, A.~Krizhevsky, J.~Ibarz, and D.~Quillen.
\newblock Learning hand-eye coordination for robotic grasping with deep
  learning and large-scale data collection.
\newblock \emph{The International Journal of Robotics Research}, pages
  173--184, 2016{\natexlab{b}}.

\bibitem[Li et~al.(2016)Li, Monroe, Ritter, Jurafsky, Galley, and
  Gao]{Li_emnlp_2016}
J.~Li, W.~Monroe, A.~Ritter, D.~Jurafsky, M.~Galley, and J.~Gao.
\newblock Deep reinforcement learning for dialogue generation.
\newblock In \emph{EMNLP}, pages 1192--1202, 2016.

\bibitem[Li et~al.(2017{\natexlab{a}})Li, Chen, Li, Gao, and
  Celikyilmaz]{li2017end}
X.~Li, Y.-N. Chen, L.~Li, J.~Gao, and A.~Celikyilmaz.
\newblock End-to-end task-completion neural dialogue systems.
\newblock \emph{arXiv preprint arXiv:1703.01008}, 2017{\natexlab{a}}.

\bibitem[Li et~al.(2017{\natexlab{b}})Li, Kiseleva, de~Rijke, and
  Grotov]{li2017towards}
Z.~Li, J.~Kiseleva, M.~de~Rijke, and A.~Grotov.
\newblock Towards learning reward functions from user interactions.
\newblock In \emph{ICTIR}, pages 941--944. ACM, 2017{\natexlab{b}}.

\bibitem[Li et~al.(2019)Li, Kiseleva, and de~Rijke]{li2019dialogue}
Z.~Li, J.~Kiseleva, and M.~de~Rijke.
\newblock Dialogue generation: From imitation learning to inverse reinforcement
  learning.
\newblock In \emph{Proceedings of the AAAI Conference on Artificial
  Intelligence}, volume~33, pages 6722--6729, 2019.

\bibitem[Li et~al.(2020)Li, Lee, Peng, Li, Shayandeh, and Gao]{li2020guided}
Z.~Li, S.~Lee, B.~Peng, J.~Li, S.~Shayandeh, and J.~Gao.
\newblock Guided dialog policy learning without adversarial learning in the
  loop.
\newblock \emph{arXiv preprint arXiv:2004.03267}, 2020.

\bibitem[Lipton et~al.(2018)Lipton, Li, Gao, Li, Ahmed, and
  Deng]{lipton2018bbq}
Z.~Lipton, X.~Li, J.~Gao, L.~Li, F.~Ahmed, and L.~Deng.
\newblock Bbq-networks: Efficient exploration in deep reinforcement learning
  for task-oriented dialogue systems.
\newblock In \emph{Thirty-Second AAAI Conference on Artificial Intelligence},
  2018.

\bibitem[Liu et~al.(2016)Liu, Lowe, Serban, Noseworthy, Charlin, and
  Pineau]{liu-etal-2016-evaluate}
C.-W. Liu, R.~Lowe, I.~Serban, M.~Noseworthy, L.~Charlin, and J.~Pineau.
\newblock How {NOT} to evaluate your dialogue system: An empirical study of
  unsupervised evaluation metrics for dialogue response generation.
\newblock In \emph{EMNLP}, 2016.

\bibitem[Lovett(2006)]{Lovett_choice_theory}
F.~Lovett.
\newblock Rational choice theory and explanation.
\newblock \emph{Rationality and Society}, 18\penalty0 (2):\penalty0 237--272,
  2006.

\bibitem[Lowe et~al.(2017)Lowe, Noseworthy, Serban, Angelard-Gontier, Bengio,
  and Pineau]{lowe2017towards}
R.~Lowe, M.~Noseworthy, I.~V. Serban, N.~Angelard-Gontier, Y.~Bengio, and
  J.~Pineau.
\newblock Towards an automatic turing test: Learning to evaluate dialogue
  responses.
\newblock In \emph{ACL}, pages 1116--1126, 2017.

\bibitem[Mitchell and Shneiderman(1989)]{mitchell1989dynamic}
J.~Mitchell and B.~Shneiderman.
\newblock Dynamic versus static menus: an exploratory comparison.
\newblock \emph{ACM SigCHI Bulletin}, 20\penalty0 (4):\penalty0 33--37, 1989.

\bibitem[Mnih et~al.(2015)Mnih, Kavukcuoglu, Silver, Rusu, Veness, Bellemare,
  Graves, Riedmiller, Fidjeland, Ostrovski, et~al.]{mnih2015human}
V.~Mnih, K.~Kavukcuoglu, D.~Silver, A.~A. Rusu, J.~Veness, M.~G. Bellemare,
  A.~Graves, M.~Riedmiller, A.~K. Fidjeland, G.~Ostrovski, et~al.
\newblock Human-level control through deep reinforcement learning.
\newblock \emph{Nature}, 518\penalty0 (7540):\penalty0 529, 2015.

\bibitem[Mohri et~al.(2012)Mohri, Rostamizadeh, and
  Talwalkar]{mohri2012foundations}
M.~Mohri, A.~Rostamizadeh, and A.~Talwalkar.
\newblock \emph{Foundations of Machine Learning}.
\newblock MIT Press, 2012.

\bibitem[Monfort et~al.(2015)Monfort, Liu, and Ziebart]{monfort_aaai_2015}
M.~Monfort, A.~Liu, and B.~Ziebart.
\newblock Intent prediction and trajectory forecasting via predictive inverse
  linear-quadratic regulation.
\newblock In \emph{AAAI}, pages 3672--3678. {AAAI} Press, 2015.

\bibitem[Ng and Russell(2000)]{ng_icml_2000}
A.~Y. Ng and S.~J. Russell.
\newblock Algorithms for inverse reinforcement learning.
\newblock In \emph{ICML}, pages 663--670. ACM, 2000.

\bibitem[Obendorf et~al.(2007)Obendorf, Weinreich, Herder, and
  Mayer]{obendorf2007web}
H.~Obendorf, H.~Weinreich, E.~Herder, and M.~Mayer.
\newblock Web page revisitation revisited: implications of a long-term
  click-stream study of browser usage.
\newblock In \emph{Proceedings of the SIGCHI conference on Human factors in
  computing systems}, pages 597--606, 2007.

\bibitem[Odijk et~al.(2015)Odijk, Meij, Sijaranamual, and
  de~Rijke]{odijk-dynamic-2015}
D.~Odijk, E.~Meij, I.~Sijaranamual, and M.~de~Rijke.
\newblock Dynamic query modeling for related content finding.
\newblock In \emph{SIGIR}, pages 33--42. ACM, 2015.

\bibitem[Papineni et~al.(2002)Papineni, Roukos, Ward, and
  Zhu]{papineni2002bleu}
K.~Papineni, S.~Roukos, T.~Ward, and W.-J. Zhu.
\newblock Bleu: a method for automatic evaluation of machine translation.
\newblock In \emph{ACL}, pages 311--318. ACL, 2002.

\bibitem[Peng et~al.(2017)Peng, Li, Li, Gao, Celikyilmaz, Lee, and
  Wong]{peng2017composite}
B.~Peng, X.~Li, L.~Li, J.~Gao, A.~Celikyilmaz, S.~Lee, and K.-F. Wong.
\newblock Composite task-completion dialogue policy learning via hierarchical
  deep reinforcement learning.
\newblock \emph{arXiv preprint arXiv:1704.03084}, 2017.

\bibitem[Peng et~al.(2018)Peng, Li, Gao, Liu, Chen, and
  Wong]{peng2018adversarial}
B.~Peng, X.~Li, J.~Gao, J.~Liu, Y.-N. Chen, and K.-F. Wong.
\newblock Adversarial advantage actor-critic model for task-completion dialogue
  policy learning.
\newblock In \emph{ICASSP}, pages 6149--6153. IEEE, 2018.

\bibitem[Perner and Lang(1999)]{perner1999development}
J.~Perner and B.~Lang.
\newblock Development of theory of mind and executive control.
\newblock \emph{Trends in Cognitive Sciences}, 3\penalty0 (9):\penalty0
  337--344, 1999.

\bibitem[Pietquin(2013)]{pietquin2013inverse}
O.~Pietquin.
\newblock Inverse reinforcement learning for interactive systems.
\newblock In \emph{Workshop on Machine Learning for Interactive Systems}, pages
  71--75. ACM, 2013.

\bibitem[Qureshi et~al.(2019)Qureshi, Boots, and Yip]{qureshi2018adversarial}
A.~H. Qureshi, B.~Boots, and M.~C. Yip.
\newblock Adversarial imitation via variational inverse reinforcement learning.
\newblock In \emph{ICLR}, 2019.

\bibitem[Ratliff et~al.(2009)Ratliff, Silver, and Bagnell]{ratliff2009learning}
N.~D. Ratliff, D.~Silver, and J.~A. Bagnell.
\newblock Learning to search: Functional gradient techniques for imitation
  learning.
\newblock \emph{Autonomous Robots}, 27\penalty0 (1):\penalty0 25--53, 2009.

\bibitem[Reddy et~al.(2019)Reddy, Dragan, Levine, Legg, and
  Leike]{reddy2019reward}
S.~Reddy, A.~D. Dragan, S.~Levine, S.~Legg, and J.~Leike.
\newblock Learning human objectives by evaluating hypothetical behavior.
\newblock \emph{arXiv preprint arXiv:1912.05652}, 2019.

\bibitem[Russell(1998)]{Russell_1998aa}
S.~Russell.
\newblock Learning agents for uncertain environments.
\newblock In \emph{COLT}, pages 101--103. ACM, 1998.

\bibitem[Saracevic(1975)]{sara:rele75}
T.~Saracevic.
\newblock Relevance: A review of and a framework for the thinking on the notion
  in information science.
\newblock \emph{Journal of the American Society for Information Science and
  Technology}, 26:\penalty0 321--343, 1975.

\bibitem[Saracevic et~al.(1988)Saracevic, Kantor, Chamis, and
  Trivison]{sara:stud88}
T.~Saracevic, P.~B. Kantor, A.~Y. Chamis, and D.~Trivison.
\newblock A study of information seeking and retrieving. {I}. background and
  methodology. {II}. users, questions and effectiveness. {III}. searchers,
  searches, overlap.
\newblock \emph{Journal of the American Society for Information Science},
  39:\penalty0 161--176; 177--196; 197--216, 1988.

\bibitem[Schnabel et~al.(2019)Schnabel, Bennett, and
  Joachims]{schnabel2019shaping}
T.~Schnabel, P.~N. Bennett, and T.~Joachims.
\newblock Shaping feedback data in recommender systems with interventions based
  on information foraging theory.
\newblock In \emph{Proceedings of the Twelfth ACM International Conference on
  Web Search and Data Mining}, pages 546--554, 2019.

\bibitem[Schrittwieser et~al.(2019)Schrittwieser, Antonoglou, Hubert, Simonyan,
  Sifre, Schmitt, Guez, Lockhart, Hassabis, Graepel,
  et~al.]{schrittwieser2019mastering}
J.~Schrittwieser, I.~Antonoglou, T.~Hubert, K.~Simonyan, L.~Sifre, S.~Schmitt,
  A.~Guez, E.~Lockhart, D.~Hassabis, T.~Graepel, et~al.
\newblock Mastering atari, go, chess and shogi by planning with a learned
  model.
\newblock \emph{arXiv preprint arXiv:1911.08265}, 2019.

\bibitem[Schulman et~al.(2017)Schulman, Wolski, Dhariwal, Radford, and
  Klimov]{schulman2017proximal}
J.~Schulman, F.~Wolski, P.~Dhariwal, A.~Radford, and O.~Klimov.
\newblock Proximal policy optimization algorithms.
\newblock \emph{arXiv preprint arXiv:1707.06347}, 2017.

\bibitem[Sepliarskaia et~al.(2018)Sepliarskaia, Kiseleva, Radlinski, and
  de~Rijke]{sepliarskaia2018preference}
A.~Sepliarskaia, J.~Kiseleva, F.~Radlinski, and M.~de~Rijke.
\newblock Preference elicitation as an optimization problem.
\newblock In \emph{Proceedings of the 12th ACM Conference on Recommender
  Systems}, pages 172--180, 2018.

\bibitem[Seyed~Ghasemipour et~al.(2019)Seyed~Ghasemipour, Gu, and
  Zemel]{NIPS2019_9002}
S.~K. Seyed~Ghasemipour, S.~S. Gu, and R.~Zemel.
\newblock Smile: Scalable meta inverse reinforcement learning through
  context-conditional policies.
\newblock In \emph{Advances in Neural Information Processing Systems 32}, pages
  7881--7891. 2019.

\bibitem[Shani et~al.(2005)Shani, Heckerman, and Brafman]{shani2005mdp}
G.~Shani, D.~Heckerman, and R.~I. Brafman.
\newblock An {MDP}-based recommender system.
\newblock \emph{Journal of Machine Learning Research}, 6\penalty0
  (Sep):\penalty0 1265--1295, 2005.

\bibitem[Silver et~al.(2016)Silver, Huang, Maddison, Guez, Sifre, Van
  Den~Driessche, Schrittwieser, Antonoglou, Panneershelvam, Lanctot,
  et~al.]{silver2016go}
D.~Silver, A.~Huang, C.~J. Maddison, A.~Guez, L.~Sifre, G.~Van Den~Driessche,
  J.~Schrittwieser, I.~Antonoglou, V.~Panneershelvam, M.~Lanctot, et~al.
\newblock Mastering the game of go with deep neural networks and tree search.
\newblock \emph{Nature}, 529\penalty0 (7587):\penalty0 484--489, 2016.

\bibitem[Su et~al.(2018)Su, Li, Gao, Liu, and Chen]{Su2018D3Q}
S.-Y. Su, X.~Li, J.~Gao, J.~Liu, and Y.-N. Chen.
\newblock Discriminative deep dyna-q: Robust planning for dialogue policy
  learning.
\newblock In \emph{EMNLP}, 2018.

\bibitem[Sutton and Barto(1998)]{sutton1998rl}
R.~S. Sutton and A.~G. Barto.
\newblock \emph{Reinforcement learning: An introduction}.
\newblock MIT press Cambridge, 1998.

\bibitem[Sutton and Barto(2018)]{sutton2018reinforcement}
R.~S. Sutton and A.~G. Barto.
\newblock \emph{Reinforcement learning: An introduction}.
\newblock MIT press, 2018.

\bibitem[Takanobu et~al.(2019)Takanobu, Zhu, and Huang]{takanobu2019guided}
R.~Takanobu, H.~Zhu, and M.~Huang.
\newblock Guided dialog policy learning: Reward estimation for multi-domain
  task-oriented dialog.
\newblock \emph{arXiv preprint arXiv:1908.10719}, 2019.

\bibitem[Teevan(2008)]{teevan2008people}
J.~Teevan.
\newblock How people recall, recognize, and reuse search results.
\newblock \emph{ACM Transactions on Information Systems (TOIS)}, 26\penalty0
  (4):\penalty0 1--27, 2008.

\bibitem[ter Hoeve et~al.(2020)ter Hoeve, Sim, Nouri, Fourney, de~Rijke, and
  White]{ter2020conversations}
M.~ter Hoeve, R.~Sim, E.~Nouri, A.~Fourney, M.~de~Rijke, and R.~W. White.
\newblock Conversations with documents: An exploration of document-centered
  assistance.
\newblock In \emph{CHIIR}, pages 43--52. ACM, 2020.

\bibitem[Varian(1999)]{varian1999economics}
H.~R. Varian.
\newblock Economics and search.
\newblock In \emph{ACM SIGIR Forum}, volume~33, pages 1--5. ACM New York, NY,
  USA, 1999.

\bibitem[Vinyals et~al.(2019)Vinyals, Babuschkin, Czarnecki, Mathieu, Dudzik,
  Chung, Choi, Powell, Ewalds, Georgiev, et~al.]{vinyals2019grandmaster}
O.~Vinyals, I.~Babuschkin, W.~M. Czarnecki, M.~Mathieu, A.~Dudzik, J.~Chung,
  D.~H. Choi, R.~Powell, T.~Ewalds, P.~Georgiev, et~al.
\newblock Grandmaster level in starcraft ii using multi-agent reinforcement
  learning.
\newblock \emph{Nature}, 575\penalty0 (7782):\penalty0 350--354, 2019.

\bibitem[Wang et~al.(2016)Wang, Kurth-Nelson, Tirumala, Soyer, Leibo, Munos,
  Blundell, Kumaran, and Botvinick]{wang2016learning}
J.~X. Wang, Z.~Kurth-Nelson, D.~Tirumala, H.~Soyer, J.~Z. Leibo, R.~Munos,
  C.~Blundell, D.~Kumaran, and M.~Botvinick.
\newblock Learning to reinforcement learn.
\newblock \emph{arXiv preprint arXiv:1611.05763}, 2016.

\bibitem[Wei et~al.(2017)Wei, Zhang, Yuan, Cao, Fu, Xie, Rui, and
  Ma]{wei_wsdm_2017}
H.~Wei, F.~Zhang, N.~J. Yuan, C.~Cao, H.~Fu, X.~Xie, Y.~Rui, and W.-Y. Ma.
\newblock Beyond the words: Predicting user personality from heterogeneous
  information.
\newblock In \emph{WSDM}, pages 305--314. ACM, 2017.

\bibitem[White(2016)]{white2016interactions}
R.~W. White.
\newblock \emph{Interactions with search systems}.
\newblock Cambridge University Press, 2016.

\bibitem[White et~al.(2002)White, Ruthven, and Jose]{white2002finding}
R.~W. White, I.~Ruthven, and J.~M. Jose.
\newblock Finding relevant documents using top ranking sentences: an evaluation
  of two alternative schemes.
\newblock In \emph{SIGIR}, pages 57--64, 2002.

\bibitem[White et~al.(2005)White, Ruthven, Jose, and van
  Rijsbergen]{white2005evaluating}
R.~W. White, I.~Ruthven, J.~M. Jose, and C.~van Rijsbergen.
\newblock Evaluating implicit feedback models using searcher simulations.
\newblock \emph{ACM Transactions on Information Systems (TOIS)}, 23\penalty0
  (3):\penalty0 325--361, 2005.

\bibitem[Williams et~al.(2017)Williams, Asadi, and Zweig]{williams2017hybrid}
J.~D. Williams, K.~Asadi, and G.~Zweig.
\newblock Hybrid code networks: practical and efficient end-to-end dialog
  control with supervised and reinforcement learning.
\newblock \emph{arXiv preprint arXiv:1702.03274}, 2017.

\bibitem[Williams et~al.(2016{\natexlab{a}})Williams, Kiseleva, Crook, Zitouni,
  Awadallah, and Khabsa]{Williams_sigir_2016}
K.~Williams, J.~Kiseleva, A.~Crook, I.~Zitouni, A.~H. Awadallah, and M.~Khabsa.
\newblock Is this your final answer? evaluating the effect of answers on good
  abandonment in mobile search.
\newblock In \emph{SIGIR}, pages 889--892, 2016{\natexlab{a}}.

\bibitem[Williams et~al.(2016{\natexlab{b}})Williams, Kiseleva, Crook, Zitouni,
  Awadallah, and Khabsa]{Williams_www_2016}
K.~Williams, J.~Kiseleva, A.~C. Crook, I.~Zitouni, A.~H. Awadallah, and
  M.~Khabsa.
\newblock Detecting good abandonment in mobile search.
\newblock In \emph{WWW}, pages 495--505, 2016{\natexlab{b}}.

\bibitem[Yilmaz et~al.(2014)Yilmaz, Verma, Craswell, Radlinski, and
  Bailey]{Yilmaz_cikm_2014}
E.~Yilmaz, M.~Verma, N.~Craswell, F.~Radlinski, and P.~Bailey.
\newblock Relevance and effort: An analysis of document utility.
\newblock In \emph{CIKM}, pages 91--100, 2014.

\bibitem[Zhang and Dragan(2019)]{zhang2019learning}
J.~Y. Zhang and A.~D. Dragan.
\newblock Learning from extrapolated corrections.
\newblock In \emph{2019 International Conference on Robotics and Automation
  (ICRA)}, pages 7034--7040. IEEE, 2019.

\bibitem[Zhao et~al.(2018)Zhao, Zhang, Ding, Xia, Tang, and
  Yin]{zhao2018recommendations}
X.~Zhao, L.~Zhang, Z.~Ding, L.~Xia, J.~Tang, and D.~Yin.
\newblock Recommendations with negative feedback via pairwise deep
  reinforcement learning.
\newblock In \emph{KDD}, pages 1040--1048, 2018.

\bibitem[Zheng et~al.(2018)Zheng, Zhang, Zheng, Xiang, Yuan, Xie, and
  Li]{zheng2018drn}
G.~Zheng, F.~Zhang, Z.~Zheng, Y.~Xiang, N.~J. Yuan, X.~Xie, and Z.~Li.
\newblock Drn: A deep reinforcement learning framework for news recommendation.
\newblock In \emph{WWW}, pages 167--176, 2018.

\bibitem[Zhu et~al.(2017)Zhu, Mottaghi, Kolve, Lim, Gupta, Fei-Fei, and
  Farhadi]{zhu2017target}
Y.~Zhu, R.~Mottaghi, E.~Kolve, J.~J. Lim, A.~Gupta, L.~Fei-Fei, and A.~Farhadi.
\newblock Target-driven visual navigation in indoor scenes using deep
  reinforcement learning.
\newblock In \emph{ICRA}, pages 3357--3364. IEEE, 2017.

\bibitem[Ziebart et~al.(2012)Ziebart, Dey, and Bagnell]{ziebart_iui_2012}
B.~Ziebart, A.~Dey, and J.~A. Bagnell.
\newblock Probabilistic pointing target prediction via inverse optimal control.
\newblock In \emph{IUI}, pages 1--10. ACM, 2012.

\bibitem[Ziebart(2010)]{ziebart2010modeling}
B.~D. Ziebart.
\newblock \emph{Modeling Purposeful Adaptive Behavior with the Principle of
  Maximum Causal Entropy}.
\newblock PhD thesis, Carnegie Mellon University, 2010.

\bibitem[Ziebart et~al.(2008)Ziebart, Maas, Bagnell, and
  Dey]{ziebart_aaai_2008}
B.~D. Ziebart, A.~L. Maas, J.~A. Bagnell, and A.~K. Dey.
\newblock Maximum entropy inverse reinforcement learning.
\newblock In \emph{AAAI}, pages 1433--1438. {AAAI} Press, 2008.

\end{thebibliography}


\end{document}